\pdfoutput=1
\documentclass[11pt]{article}

\usepackage[preprint]{ACL/acl}

\usepackage{times}
\usepackage{latexsym}

\usepackage[T1]{fontenc}
\usepackage{float}
\usepackage[utf8]{inputenc}

\usepackage{microtype}

\usepackage{inconsolata}

\usepackage{graphicx}

\usepackage{amsmath,amsfonts,amssymb,amsthm,mathtools}
\usepackage{bm}
\usepackage{bbm}
\usepackage{longtable}
\usepackage{todonotes}
\usepackage{multirow}
\usepackage{booktabs}
\usepackage{makecell}
\usepackage{ragged2e}

\usepackage{amsmath,amsfonts,bm}

\def\eqref#1{equation~\ref{#1}}
\def\1{\bm{1}}

\def\vtheta{{\bm{\theta}}}

\def\vc{{\bm{c}}}

\def\vt{{\bm{t}}}

\def\vx{{\bm{x}}}
\def\vy{{\bm{y}}}

\DeclareMathAlphabet{\mathsfit}{\encodingdefault}{\sfdefault}{m}{sl}
\SetMathAlphabet{\mathsfit}{bold}{\encodingdefault}{\sfdefault}{bx}{n}

\def\vtheta{{\bm{\theta}}}

\def\vc{{\bm{c}}}

\def\vt{{\bm{t}}}

\def\vx{{\bm{x}}}
\def\vy{{\bm{y}}}

\def\vvarepsilon{{\ensuremath{\bm{\varepsilon}}}}
\def\vtheta{{\bm{\theta}}}

\definecolor{blue}{HTML}{1f77b4}
\definecolor{orange}{HTML}{ff7f0e}

\def\BModel{{\ensuremath{\textcolor{blue}{\bar{M}}}}}
\def\OModel{{\ensuremath{\textcolor{orange}{\widetilde{M}}}}}

\def\oepsilon#1{%
  \ensuremath{
    \textcolor{orange}{\widetilde{\vvarepsilon} %
    \ifx\relax#1\relax\else_{#1}\fi %
    } %
  }%
}

\def\bepsilon#1{%
  \ensuremath{
    \textcolor{blue}{\bar{\vvarepsilon}%
    \ifx\relax#1\relax\else_{#1}\fi%
    }%
  }%
}

\def\oobject#1{%
  \ensuremath{
    \textcolor{orange}{\widetilde{o} %
    \ifx\relax#1\relax\else_{#1}\fi %
    } %
  }%
}

\def\bobject#1{%
  \ensuremath{
    \textcolor{blue}{\bar{o} %
    \ifx\relax#1\relax\else_{#1}\fi %
    } %
  }%
}

\def\osubject#1{%
  \ensuremath{
    \textcolor{orange}{\widetilde{s} %
    \ifx\relax#1\relax\else_{#1}\fi %
    } %
  }%
}

\def\bsubject#1{%
  \ensuremath{
    \textcolor{blue}{\bar{s} %
    \ifx\relax#1\relax\else_{#1}\fi %
    } %
  }%
}

\def\orelation#1{%
  \ensuremath{
    \textcolor{orange}{r %
    \ifx\relax#1\relax\else_{#1}\fi %
    } %
  }%
}

\def\brelation#1{%
  \ensuremath{
    \textcolor{blue}{r%
    \ifx\relax#1\relax\else_{#1}\fi%
    }%
  }%
}

\def\cityEpsilon{%
  \ensuremath{
    \textcolor{orange}{\vvarepsilon_{\texttt{cityOf}} %
    } %
  }%
}

\def\capitalEpsilon{%
  \ensuremath{
    \textcolor{blue}{\vvarepsilon_{\texttt{capitalOf}} %
    } %
  }%
}

\usepackage{tabularx}
\usepackage{booktabs}
\usepackage{xcolor,colortbl}
\usepackage{xspace}
\usepackage{multirow}
\usepackage{array}
\newcolumntype{P}[1]{>{\centering\arraybackslash}p{#1}}
\newcolumntype{M}[1]{>{\centering\arraybackslash}m{#1}}
\usepackage[export]{adjustbox}

\usepackage{listings}
\newcommand{\methodname}{\textsc{Causal Diagnosticity}\xspace}

\usepackage[most]{tcolorbox}
\colorlet{yellow}{yellow!40!}

\newtcbox\tcbpink{
        on line,
        boxsep=4pt, left=0pt,right=0pt,top=0pt,bottom=0pt,
        colframe=white,colback=pink!40!,  
        highlight math style={enhanced}
        }

\newtcbox\tcbblue{
        on line,
        boxsep=4pt, left=0pt,right=0pt,top=0pt,bottom=0pt,
        colframe=white,colback=blue!40!,  
        highlight math style={enhanced}
        }

\definecolor{forestgreen}{rgb}{0.13, 0.55, 0.13}

\usepackage{hyperref}
\usepackage{url}

\title{A Causal Lens for Evaluating Faithfulness Metrics}
\author{Kerem Zaman \quad Shashank Srivastava\\
  UNC Chapel Hill \\
  \texttt{\{kzaman, ssrivastava\}@cs.unc.edu}
  }

\begin{document}
\maketitle
\begin{abstract}
Large Language Models (LLMs) offer natural language explanations as an alternative to feature attribution methods for model interpretability. However, despite their plausibility, they may not reflect the model's true reasoning faithfully. While several faithfulness metrics have been proposed, they are often evaluated in isolation, making principled comparisons between them difficult. We present \methodname, a testbed framework for evaluating faithfulness metrics for natural language explanations. We use the concept of diagnosticity, and employ model-editing methods to generate faithful-unfaithful explanation pairs. Our benchmark includes four tasks: fact-checking, analogy, object counting, and multi-hop reasoning. We evaluate prominent faithfulness metrics, including post-hoc explanation and chain-of-thought methods. Diagnostic performance varies across tasks and models, with Filler Tokens performing best overall. Additionally, continuous metrics are generally more diagnostic than binary ones but can be sensitive to noise and model choice. Our results highlight the need for more robust faithfulness metrics. \footnote{Code available at \url{https://github.com/KeremZaman/CausalDiagnosticity}}
\end{abstract}

\section{Introduction}
\label{sec:introduction}

Natural language explanations from Large Language Models (LLMs) have enhanced possibilities for model interpretability, offering readable insights that surpass traditional feature attribution methods. Most LLMs can generate explanations for their predictions at minimal cost~\cite{Wei2022ChainOT}. However, despite fluency and plausibility, such explanations may not reflect the model's actual reasoning process and can mislead practitioners~\cite{Turpin2023LanguageMD}.

The idea of \textit{faithfulness} aims to assess how accurately explanations reflect the true reasoning mechanism of the model. While numerous metrics have been proposed to measure faithfulness for natural language-based explanations, the field lacks a principled framework for evaluating these metrics themselves. We cannot trust a faithfulness metric if we cannot reliably assess whether it actually distinguishes faithful from unfaithful explanations. \citet{Parcalabescu2023OnMF} made initial progress by comparing different metrics on the same data and models, yet their work did not evaluate the effectiveness of the metrics themselves.

We address this critical gap through \methodname, an evaluation framework for rigorously comparing faithfulness metrics. Our framework operationalizes the concept of \textit{diagnosticity}~\cite{Chan2022ACS}, which measures how often a faithfulness metric correctly favors faithful explanations over unfaithful ones. We extend it to natural language explanations through a causal intervention approach. Rather than relying on heuristically generated unfaithful explanations, we leverage knowledge editing techniques to causally generate controlled pairs of faithful and unfaithful explanations, ensuring ground truth for evaluation.

\begin{figure}
    \centering
    \includegraphics[width=\linewidth]{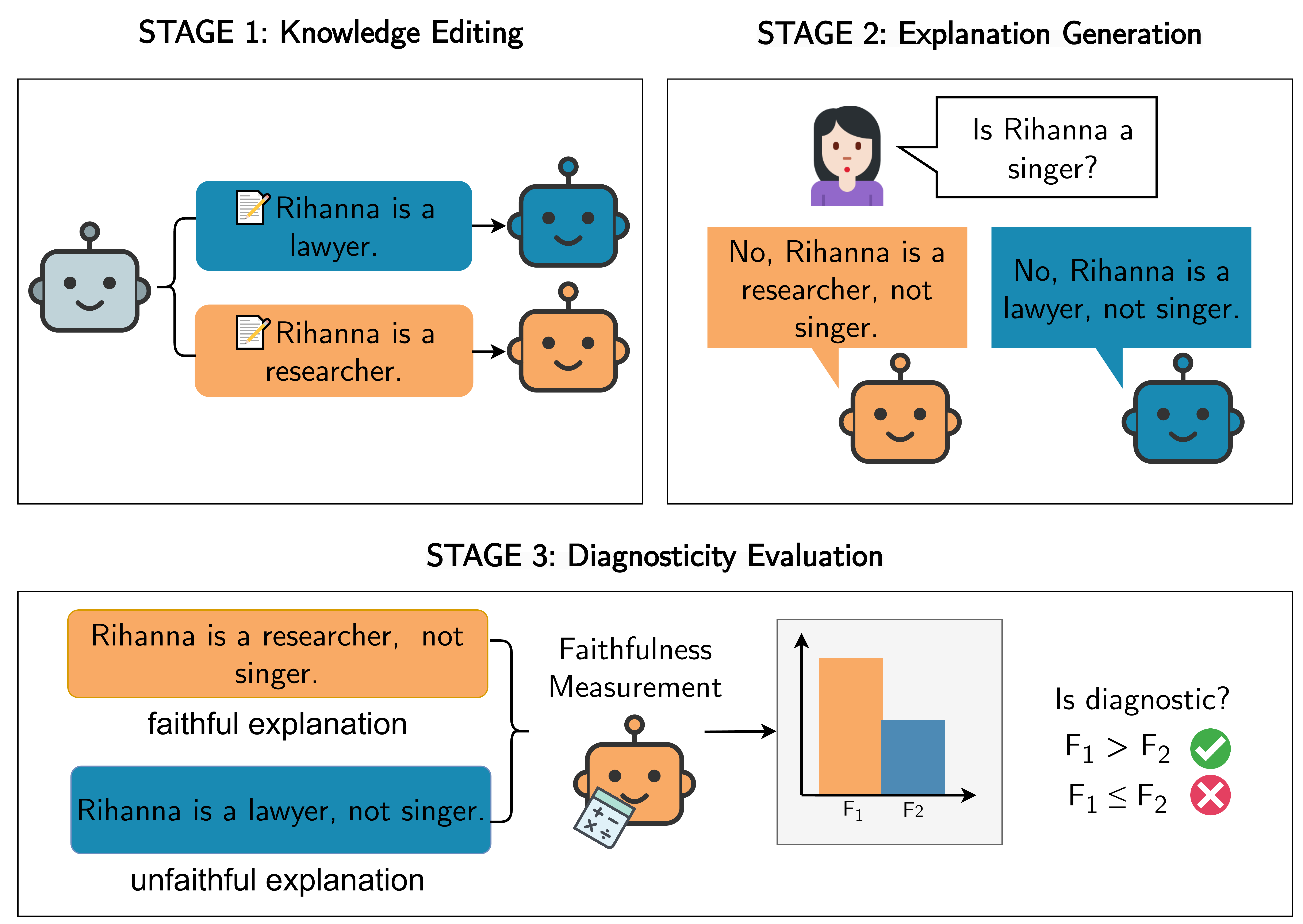}
    \caption{Our framework has three stages: (1) \textbf{Knowledge Editing}: applying counterfactual edits to models; (2) \textbf{Explanation Generation}: generating faithful and unfaithful explanation pairs using the edited models, or synthetically generating such pairs based on the edits; (3) \textbf{Diagnosticity Evaluation}: assessing the chosen faithfulness metric with one of the edited models using the faithful-unfaithful explanation pairs. Diagnostic faithfulness metrics should assign a higher score to the faithful explanations than to the unfaithful ones.}
    \label{fig:method_overview}
\end{figure}

Our framework consists of four tasks spanning complexity levels: (1) fact-checking, (2) analogy completion, (3) object counting, and (4) multi-hop reasoning. Figure~\ref{fig:method_overview} illustrates our approach. Using this benchmark, we conduct the first systematic evaluation of prominent faithfulness metrics, including Simulatability, corruption-based Chain-of-Thought (CoT) metrics~\cite{Lanham2023MeasuringFI}, and CC-SHAP~\citep{Parcalabescu2023OnMF}.

Our findings show that the most diagnostic metric varies by task and model, but the \textbf{Filler Tokens metric emerges as the most reliable overall}. We also note that \textbf{metrics producing continuous scores are more diagnostic than those with binary scores}. That said, continuous metrics can be overly sensitive to noise and model choice. Thus, we need \textbf{more robust faithfulness metrics} that exhibit consistent behavior across models and tasks. 

Our contributions are (1) a framework for evaluating faithfulness metrics for natural language explanations, (2) a dataset spanning four tasks, and (3) benchmarking of prominent faithfulness metrics across knowledge editing methods and models to provide insights into their reliability.

\section{Technical Background}
\label{sec:background}
\paragraph{Faithfulness}  We adopt the commonly used definition of faithfulness, which is the extent to which an explanation accurately reflects the reasoning process behind a model’s prediction, following \citet{Jacovi2020TowardsFI}. While this notion is widely accepted, its concrete instantiations vary depending on the type of explanation and the method used to measure faithfulness. For example, post-hoc explanations typically capture reasoning over an input–output pair, whereas CoT explanations represent reasoning generated from the input alone. Metrics differ in how they operationalize faithfulness: some evaluate the change in predictions and explanations after modifying the input \citep{Turpin2023LanguageMD, Atanasova2023FaithfulnessTF, siegel-etal-2024-probabilities}, while others assess the change in prediction after corrupting the explanation \citep{Lanham2023MeasuringFI}. \citet{Parcalabescu2023OnMF} critique such metrics for relying on overly simplistic consistency measures and instead propose measuring faithfulness by comparing the contributions of the input and the explanation to the model’s prediction. In line with this criticism, \citet{Tutek2025MeasuringCO} intervene in model internals to measure faithfulness more directly. Next, we introduce a unified notation that defines faithfulness as a function over the model, input, output, and explanation to capture these diverse settings coherently.

Let $M_{\vtheta}$ denote an LLM parameterized by $\vtheta$ and with a context $\vc$, operating on a token set $\mathcal{V}$ such that $M(\vt^{\text{in}} \mid \vc) = \vt^{\text{out}}$, where $\vt^{\text{in}} = \langle t^{\text{in}}_1 \ldots, t^{\text{in}}_{N_{\text{in}}} \rangle$, $\vt^{\text{out}} = \langle t^{\text{out}}_1 \ldots, t_{N_{\text{out}}} \rangle$ and $\vc = \langle c_1 \ldots, c_{N_c} \rangle$ ; $t^{\text{in}}_i, t^{\text{out}}_i, c_i \in \mathcal{V}$; $N_{\text{in}}$,  $N_{\text{out}}$ and $N_c$ represent the lengths of the input, output and context sequences. The context $\vc$ consists of instructions or prompts. For brevity, we use $M$ to denote a model parameterized by $\vtheta$ with context $\vc$. The input and output sequences can take many forms. For the simplest case $\vt^{\text{in}} = \vx$ and $\vt^{\text{out}} = \vy$ where $( \vx, \vy)$ is an input-output pair for a task. With appropriate prompting, the output can take the form  $\vt^{\text{out}} = \vy \oplus \vvarepsilon $  for post-hoc explanations or $\vt^{\text{out}} = \vvarepsilon \oplus \vy $ for chain-of-thought (CoT) explanations, where $\vvarepsilon$ is the explanation and $\oplus$ denotes sequence concatenation. In our particular setup, we obtain $\vy$ from the next-token logits by selecting the token with the highest score among those corresponding to the task-specific single-token labels. We define a faithfulness metric as $\mathcal{F}\left(\vx, \vy, \vvarepsilon, M\right) \in \mathbb{R}$, where $\mathcal{F}$ represents how faithfully explanation $\vvarepsilon$ represents the reasoning process for input-output pair $(\vx,\vy)$ for the model $M$. 

\subsection{Faithfulness Metrics}

We focus on six prominent faithfulness metrics: (1)~Simulatability, metrics based on CoT corruptions~\citep{Lanham2023MeasuringFI} (including (2)~Early Answering, (3)~Adding Mistakes, (4)~Paraphrasing, and (5)~Filler Tokens), and (6)~CC-SHAP~\citep{Parcalabescu2023OnMF}. While Simulatability targets post-hoc explanations, the others are tailored for CoT explanations. CC-SHAP is applicable to both types of explanations. Next, we briefly summarize three broad categories of these metrics.

\paragraph{Simulatability} Simulatability assesses faithfulness from the lens of the extent to which whether an explanation enables a simulator to predict the model's output  \citep{DoshiVelez2017TowardsAR, Hase2020EvaluatingEA, Hase2020LeakageAdjustedSC, Wiegreffe2020MeasuringAB, Chan2022FRAMEER}. We follow \citet{Chan2022FRAMEER}'s definition of simulatability as $\mathbbm{1}_S (\vy_i \mid \vx_i, \vvarepsilon_i) - \mathbbm{1}_S (\vy_i \mid \vx_i)$, where $\mathbbm{1}_S(b \mid a)$ is the accuracy of $S$ in predicting $b$ given $a$. We use a smaller LLM as simulator in our experiments.

\paragraph{Corrupting CoT} \citet{Lanham2023MeasuringFI} identify four corruption techniques to measure CoT-faithfulness: (1) \textit{Early Answering}, truncating the CoT to get an early answer; (2) \textit{Adding Mistakes}, introducing mistakes into the CoT, and regenerating; (3) \textit{Paraphrasing}, paraphrasing the CoT and regenerating; and (4) \textit{Filler Tokens}, replacing the CoT with ellipses. An explanation is considered unfaithful if the corruption does not alter the original prediction (except for Paraphrasing, where prediction changes signify unfaithfulness). While these metrics were  originally proposed as binary measures, we extend them to quantify faithfulness as the change in prediction score for the top-predicted class before and after corruption, denoted $\hat{z}_i$ and $\hat{z}^{\prime}_i$, respectively. The faithfulness score is computed as $(\hat{z}_i - \hat{z}^{\prime}_i)$, where a larger drop following corruption indicates a more faithful explanation. For Paraphrasing, we reverse this definition and use $1 - (\hat{z}_i - \hat{z}^{\prime}_i)$. \footnote{We note that the choice between the signed difference and the unsigned difference ($|\hat{z}_i - \hat{z}^{\prime}_i|$) has a significant impact on the results. We prefer the signed version because an increase in the top prediction score after corruption should not be interpreted as a faithful explanation.}

\paragraph{CC-SHAP} \citet{Parcalabescu2023OnMF} assess faithfulness by aligning input contributions to prediction and explanation using SHAP \citep{Lundberg2017AUA} scores. They calculate importance scores for each input token's prediction, then for each token in the explanation, aggregating them. Convergence of these score distributions is then measured. This method is applicable to both post-hoc and Chain-of-Thought (CoT) explanations. We describe each metric in more detail in Appendix \ref{appendix:metrics}.

\subsection{Knowledge Editing}
Our framework generates faithful-unfaithful explanation pairs by modifying facts within LLMs using knowledge editing.
Knowledge editing methods allow updates without altering unrelated knowledge \citep{cohen2024evaluating,Zhang2024ACS, Patil2023CanSI, geva-etal-2023-dissecting, gupta-etal-2023-editing, hartvigsen2023aging,zheng-etal-2023-edit, Meng2022LocatingAE}, using triplets of subjects \textit{s}, objects \textit{o}, and relations \textit{r}. For instance, they can update ($s = \text{Joe Biden}$, $r = \text{is the president of}$, $o = \text{the United States}$) to ($s = \text{Donald Trump}$, $r = \text{is the president of}$, $o = \text{the United States}$). We explore two knowledge editing methods: (1) In-Context Editing (ICE) \citep{Cohen2023EvaluatingTR}, and (2) MEMIT \citep{meng2023memit}. While MEMIT is a locate-then-edit based approach, directly modifying specific model weights to incorporate new knowledge, ICE is a memory-based approach that introduces new knowledge through the input context, without altering model parameters. Unlike ICE, MEMIT relies on a rigid subject-object-target template, which limits its use in complex scenarios. Additionally, MEMIT-like methods are highly sensitive to hyperparameters requiring model-specific tuning or limiting use to models with known optimal values \citep{Wang2023EasyEditAE}. Finally, in-context approaches consistently surpass MEMIT in multi-step reasoning tasks~\cite{Cohen2023EvaluatingTR}. We therefore adopt ICE as our primary knowledge-editing method, but also report results for both methods as part of our ablation study (\S 5.4) to confirm that our conclusions hold across editing paradigms.

\section{Method}
\label{sec:method}
Our \methodname framework is inspired by the idea of \textit{diagnosticity}, which evaluates how well faithfulness metrics distinguish between faithful and unfaithful explanations. In \ref{subsec:diagnosticity}, we summarize diagnosticity as introduced by \citet{Chan2022ACS} for evaluating feature attribution methods. In \ref{subsec:method}, we introduce \methodname, describing how it builds on diagnosticity and extends it to natural language explanations using causal interventions via edited models.

\subsection{Diagnosticity}
\label{subsec:diagnosticity}
Despite the plethora of faithfulness metrics for natural language explanations \citep{Jacovi2020TowardsFI, Lyu2022TowardsFM}, the community lacks an evaluation framework to compare them. We adopt \textit{diagnosticity}~\cite{Chan2022ACS}, which measures how often a faithfulness metric prefers faithful over unfaithful explanations. For example, if a model correctly answers ``No" to the question, ``Is Rihanna a researcher?" based on her being a singer, a faithful explanation should reflect this reasoning. An explanation that provides an irrelevant rationale (e.g., albums she has sold) or false information (e.g., the wrong occupation) would be unfaithful.

Following the notation from \citet{Chan2022ACS}, let $u$ and $v$ be explanations (regardless of form, e.g., language, feature attributions, etc.), with $u \succ v$ denoting that $u$ is more faithful than $v$. A faithfulness metric $\mathcal{F}$ ranks the explanations as $u \succ_{\mathcal{F}} v$ if it assigns a higher faithfulness score to $u$ than $v$. Then, the diagnosticity of the metric $\mathcal{F}$ is:

\begin{equation}
D(\mathcal{F}) = P(u \succ_{\mathcal{F}} v | u \succ v)
\end{equation}

We approximate this using an empirical estimate of the probability from pairs of faithful and unfaithful explanations. Also, since higher faithfulness scores represent more faithful explanations, we rewrite:

\begin{equation}
D(\mathcal{F}) \approx \frac{1}{|Z|} \sum_{(u_i, v_i) \in Z} 
\mathbbm{1} (\mathcal{F}_{p_i, M}\left(u_i \right) > \mathcal{F}_{ p_i, M}\left(v_i\right))
\end{equation}

\noindent where $Z$ contains pairs of faithful $(u_i)$ and unfaithful $(v_i)$ explanations for input-output pairs $p_i:= (\vx_i, \vy_i)$. 

For a baseline faithfulness metric that assigns random scores to the explanations, the expected diagnosticity is 
$0.5$. To capture this behavior and relax the strict inequality, we modify the diagnosticity definition as follows:
\begin{equation}
    D(\mathcal{F}) \approx \frac{1}{|Z|} \sum_{(u_i, v_i) \in Z} d(u_i, v_i, \mathcal{F}_{p_i, M})
\end{equation}
with the pairwise function $d(\cdot)$ defined as
\begin{equation}
\small
d(u_i, v_i, \mathcal{F}_{p_i, M}) = 
\begin{cases}
  1 & \text{if } \mathcal{F}_{p_i, M}\left(u_i \right) > \mathcal{F}_{ p_i, M}\left(v_i\right) \\
  0.5 & \text{if } \mathcal{F}_{p_i, M}\left(u_i \right) = \mathcal{F}_{ p_i, M}\left(v_i\right) \\
  0 & \text{if } \mathcal{F}_{p_i, M}\left(u_i \right) < \mathcal{F}_{ p_i, M}\left(v_i\right)
\end{cases}
\end{equation}
This revised formulation accommodates the scenario where random scoring yields an expected diagnosticity of  $0.5$, by assigning a neutral score when the faithfulness scores are equal.

\subsection{Causal Diagnosticity}
\label{subsec:method}
To obtain unfaithful explanations for measuring diagnosticity, \citet{Chan2022ACS} use random feature attribution scores. While random scores can work for structured explanations like feature attributions, this approach is not straightforward for natural language explanations. Random text cannot function as a meaningful explanation and cannot ensure unfaithfulness in a coherent way. To address this, we introduce \methodname, which generates unfaithful explanations using knowledge editing. Rather than injecting random noise, we modify a model’s internal knowledge. For example, consider the \texttt{capitalOf} relation for the question ``Is Paris the capital of France?" and a model that correctly associates this to the knowledge $(s = \text{Paris}, r = \text{is the capital of}, o = \text{France})$. By altering the model's knowledge, we create two variations where the subject $s$ is replaced with Berlin or London. Both modified models should answer ``No" to the original question but for different reasons: ``No, because Berlin is the capital of France." and ``No, because London is the capital of France." In particular, each of these two explanations should be unfaithful to the model that generated the other.

Formally, let $\vy_i$ be the prediction for the input $\vx_i$ while \BModel{} and \OModel{} be the altered models. \BModel{} generates the explanation \bepsilon{i} and \OModel{} generates the explanation \oepsilon{i}. We modify diagnosticity as:

\begin{equation}
\label{eq:causal_diagnosticity}
    D(\mathcal{F}) = \frac{1}{|Z|} \sum_{(\bepsilon{i}, \oepsilon{i}) \in Z} d(\bepsilon{i}, \oepsilon{i}, \mathcal{F}_{p_i, \BModel})
\end{equation}
where
\begin{equation}
\small
d(u_i, v_i, \mathcal{F}_{p_i, \BModel}) = 
\begin{cases}
  1 & \text{if } \mathcal{F}_{p_i, \BModel}\left(\bepsilon{i} \right) > \mathcal{F}_{ p_i, \BModel}\left(\oepsilon{i}\right) \\
  0.5 & \text{if } \mathcal{F}_{p_i, \BModel}\left(\bepsilon{i} \right) = \mathcal{F}_{ p_i, \BModel}\left(\oepsilon{i}\right) \\
  0 & \text{if } \mathcal{F}_{p_i, \BModel}\left(\bepsilon{i} \right) < \mathcal{F}_{ p_i, \BModel}\left(\oepsilon{i}\right)
\end{cases}
\end{equation}

Models \BModel{} and \OModel{} are edited such that \bepsilon{i} is faithful to \BModel{}, while \oepsilon{i} is unfaithful to \BModel{}. They are created by modifying parameters $\vtheta$ or context $\vc$, depending on the knowledge editing method.  The choice of models is flexible: in most cases, either model can be used in Equation \ref{eq:causal_diagnosticity} by swapping 
$\bepsilon{i}$ and $\oepsilon{i}$. However, in some tasks, one explanation may be faithful to both models, restricting arbitrary model selection. For example, in our Analogy task (see Figure \ref{fig:tasks}), the \texttt{capitalOf} relation exists in only one model, while the \texttt{cityOf} relation holds in both. Additionally, the original model $\vtheta$ can be used as long as it satisfies the faithfulness relations with the explanation pairs. However, we created \textit{two} edited variants to guarantee that all conditions are met.

\section{Tasks}
\label{sec:tasks}
\begin{figure*}[t]
    \centering
    \includegraphics[width=\linewidth]{figures/All-Tasks-horizontal.pdf}
    \caption{Overview of the four tasks, illustrated with example questions, answers, and explanations from the edited models. The explanations can be model generated or synthetically constructed to align with specific edits. The blue and orange robots represent models $\BModel$ and $\OModel$, respectively, while the color-matched boxes indicate counterfactual knowledge injected through editing. Speech bubbles next to each model display the answer ($\vy$) and explanation ($\bepsilon{}$ or $\oepsilon{}$). Although both models generate the same answer, their reasoning differs, as reflected in the explanations.
    }
    
    \label{fig:tasks}
\end{figure*}
We evaluate faithfulness metrics using four controlled tasks in the \methodname framework: (1) fact-checking, (2) analogy, (3)  object counting, and (4) multi-hop reasoning. These tasks assess causal diagnosticity by using counterfactual models with faithful and unfaithful explanations. They are deliberately designed to span varying levels of complexity. The FactCheck task is the simplest, requiring models to answer yes/no questions with minimal reasoning. In contrast, the Analogy task introduces additional complexity through its multiple-choice format and hierarchical, non-mutually exclusive edited relations (see \S~\ref{subsec:method}). The Object Counting task, also a multiple-choice format, goes beyond simple classification by requiring models to demonstrate counting capabilities. Finally, the Multi-hop Reasoning task is the most complex, requiring multiple reasoning steps to arrive at the correct answer, and the most challenging in terms of diagnosticity, as faithful and unfaithful explanation pairs often share significant internal and lexical reasoning components. While the altered models should reason differently, their explanations may not always reference the modifications. To ensure valid faithfulness comparisons, we synthetically generate explanations that emphasize model differences. While this reduces the realism of the explanations, it is necessary to guarantee the validity of our faithful/unfaithful labels. We later analyze the impact of using synthetic vs. model-generated explanations in \S\ref{subsec:explanation_type}. Figure~\ref{fig:tasks} provides an overview of these tasks, including example inputs, outputs, and explanations.

\subsection{FactCheck Task}
\paragraph{Task} This task focuses on simple fact-checking, where a fact is presented alongside two counterfactual answers. For any relation $(s_i, r_i, o_i)$, we present a question that checks its correctness, accompanied by two counterfactuals: $(s_i, r_i, \bobject{i})$ and $(s_i, r_i, \oobject{i})$. These counterfactuals yield the same answer but are based on different reasoning. For instance, given the knowledge triplet $(s_i = \text{"Rihanna"}, r_i = \text{"is"}, o_i = \text{"a singer"})$, the corresponding question would be "Is Rihanna a singer?" Let the counterfactual objects be $\bobject{i} = \text{"researcher"}$ and $\oobject{i} = \text{"lawyer"}$. Both counterfactuals would result in the answer "No," but for different reasons.
\paragraph{Dataset} We construct our dataset from \textsc{CounterFact}  \citep{Meng2022LocatingAE}, which consists of knowledge triplets. We 
convert these triplets to yes/no questions. Then, for each object $o_i$, we fetch sibling entities from WikiData to create counterfactuals. Finally, we generate synthetic explanations corresponding to each counterfactual. For example, the explanation \bepsilon{i} could be "Joe Biden is a researcher, not the president of the United States" for \bobject{i}. See Appendix~\ref{appendix:dataset} for details.

\subsection{Analogy Task}
\paragraph{Task} This task is based on analogies exploiting hierarchies between two relations where $\brelation{1} \subset \orelation{2}$ holds. For any $(s_i, o_i)$ and $(s_j, o_j)$, there exist $(s_i, \brelation{1}, o_i)$ and $(s_j, \orelation{2}, o_j)$ such that $\brelation{1} \subset \orelation{2}$. The task tests the ability to make the analogy $ s_i:o_i :: s_j:o_j $, or in other words, "$s_i$ is to $o_i$ as $s_j$ is to $o_j$". We choose \brelation{1} and \orelation{2} as \brelation{\texttt{capitalOf}} and \orelation{\texttt{cityOf}} relations, respectively. For instance, we test "Paris is to France as Berlin is to Germany." We corrupt one of the models so that the relation \brelation{\texttt{capitalOf}} is no longer valid while the relation \orelation{\texttt{cityOf}} still holds. Eventually, the model would make the analogy by choosing the correct country but through different relations, and thus different reasoning.\\
\paragraph{Dataset} We collect a list of countries and cities, then select one capital and one non-capital city for each country. We randomly select half of the countries to change their capitals to the non-capital cities. Then, we sample 1,000 pairs, each with one country having an unchanged capital and one with a changed capital. Finally, we generate fill-in-the-blank multiple-choice questions based on these pairs, such as "Fill in the blank: Athens is to Greece like Paris is to \_\_ (A) Tonga (B) France." In this example, both \orelation{\texttt{cityOf}} and \brelation{\texttt{capitalOf}} relations provide sufficient reasoning to answer "France". While the corresponding synthetic explanation, \capitalEpsilon{}, for the model with unaltered capitals would be "The capital of France is Paris, as the capital of Greece is Athens.", the one for the model with altered capitals, \cityEpsilon{}, would be "Paris is a city in France, as Athens is a city in Greece."

\subsection{Object Counting Tasks}
\paragraph{Task} Adapted from BIG-bench~\citep{srivastava2023beyond}, this task tests object classification and counting. The model identifies how many objects in a list belong to a given category. We alter the model’s internal knowledge, swapping objects across categories while keeping the answer numerically identical but reasoning distinct. For example, in \textit{How many of "countertop," "grape," and "kiwifruit" are fruits?}, the correct answer is 2, since “countertop” is not a fruit. If the model is edited to classify “countertop” as a fruit and “grape” as furniture, the answer remains 2, but for different reasons.
\paragraph{Dataset} We define five object categories, each with five types. For each type, we collect 10 representative entities from WikiData, reserving 20\% for reassignment after model editing. We generate 1000 questions, equally split between two types: yes/no questions, asking if all or any items in a list belong to a given type, and number questions, asking how many items belong to a specific type. For both types, we randomly determine the number of items (3 to 6) and select a target type. For yes/no questions, we ensure that after knowledge editing, the number of entities of the target type remains unchanged. For number questions, we reassign one entity from the target type and one from other types. Dataset details are in Appendix~\ref{appendix:dataset}.

\subsection{Multi-hop Reasoning Task}
\paragraph{Task} This extends diagnostic evaluation to complex multi-step reasoning. Like FactCheck, it ensures identical answers across counterfactual settings but requires multi-hop chains to reach conclusions. Unlike FactCheck, it requires explanations grounded in multi-step reasoning chains.  
\paragraph{Dataset} We construct this using StrategyQA \citep{Geva2021DidAU}, a multi-hop QA benchmark that provides fact decompositions for each example. We generate two counterfactual variants for one fact per question, preserving the answer while altering the reasoning. When facts are interdependent, we propagate modifications to ensure consistency. Next, we generate explanations for each counterfactual set using the original decompositions. We use \texttt{gpt-4o} for generating counterfactuals and explanations, which we manually verify for coherence. The data consists of 200 high-quality examples.

\section{Experiments}
\label{sec:experiments}
In this section, we present our results and analyses for a series of experiments. These consist of: (1) evaluating diagnosticity scores for post-hoc and CoT-based metrics,  (2) analyzing the sensitivity of CoT-based metrics to different input corruption schemes (3) analyzing the reliability of knowledge edits, (4) studying the effect of replacing ICE with MEMIT, (5) assessing model-generated vs. synthetic explanations, and (6) comparing binary and continuous metrics. We also include an analysis about the effect of model size in Appendix \ref{appendix:model_size}.

\subsection{Diagnosticity of Faithfulness Metrics}
\begin{table*}
        \centering
        \resizebox{\linewidth}{!}{
        \begin{tabular}{clccccccccc}
        \toprule
        & \multirow{2}{*}{\textbf{Metric}} & \multicolumn{2}{c}{\textbf{FactCheck}} & \multicolumn{2}{c}{\textbf{Analogy}} & \multicolumn{2}{c}{\textbf{Object Counting}} & \multicolumn{2}{c}{\textbf{Multi-hop}} & \multirow{2}{*}{\textbf{\makecell{Copeland \\ Score ($\uparrow$)}}} \\
        \cmidrule(lr){3-4} \cmidrule(lr){5-6} \cmidrule(lr){7-8} \cmidrule(lr){9-10} &
        & \textbf{Qwen} & \textbf{Gemma} 
        & \textbf{Qwen} & \textbf{Gemma} 
        & \textbf{Qwen} & \textbf{Gemma}
        & \textbf{Qwen} & \textbf{Gemma}
        & \\
        \midrule

        \multirow{2}{*}[0.2em]{\rotatebox{90}{\textbf{p.h.}}} 
     & CC-SHAP & \underline{\textbf{0.554}} & \underline{\textbf{0.540}} & 0.345 & \underline{\textbf{0.898}} & \underline{\textbf{0.551}} & 0.466 & 0.438 & \underline{\textbf{0.658}} & \textbf{5}  \\
             & Simulatability & 0.501 & 0.507 & \textbf{0.501} & 0.501 & 0.499 & \textbf{0.500} & \textbf{0.502} & 0.512 & 3  \\
                \midrule

        \multirow{5}{*}{\rotatebox{90}{\textbf{CoT}}}
     & Early Answering & \underline{0.756} & \underline{0.838} & \underline{0.534} & \underline{0.859} & \underline{0.566} & \underline{0.724} & 0.468 & 0.435 & 18  \\
             & Filler Tokens & \underline{\textbf{0.828}} & \underline{\textbf{0.893}} & \underline{0.561} & \underline{0.810} & \underline{\textbf{0.630}} & \underline{\textbf{0.843}} & \underline{\textbf{0.682}} & \underline{\textbf{0.585}} & \textbf{29}  \\
             & Adding Mistakes & \underline{0.534} & 0.427 & \underline{\textbf{0.590}} & \underline{0.639} & \underline{0.614} & \underline{0.579} & 0.542 & 0.402 & 13  \\
             & Paraphrasing & \underline{0.556} & 0.525 & \underline{0.535} & 0.430 & 0.425 & 0.385 & 0.448 & 0.525 & 8  \\
             & CC-SHAP & \underline{0.559} & \underline{0.598} & 0.318 & \underline{\textbf{0.939}} & \underline{0.539} & 0.506 & 0.442 & 0.488 & 12  \\
                \bottomrule
        \end{tabular}
        }
        \caption{The diagnosticity scores of each metric across four tasks and two models. \textbf{Qwen} and \textbf{Gemma} correspond to \texttt{qwen2.5-7b} and \texttt{gemma-2-9b-it}, respectively. Bold numbers indicate the highest scores for each model on each task across the two categories of faithfulness metrics: post-hoc and CoT. Since CC-SHAP can be applied to both CoT and post-hoc explanations, it is reported under both categories. Underlined numbers show the diagnosticity scores that are significantly higher than $0.5$ (one-sample t-test, $p < 0.05$).}
        \label{tab:diagnosticity}
    \end{table*}

\paragraph{Experimental Setup} We evaluate the metrics described in \S \ref{sec:background} with two LLMs: \texttt{qwen-2.5-7b} \citep{Yang2024Qwen25TR}, and \texttt{gemma-2-9b-it} \citep{Riviere2024Gemma2I}. For our main experiments, we use ICE as the knowledge editing method and synthetic explanations to ensure faithfulness to the edited model.

Table \ref{tab:diagnosticity} reports diagnosticity scores across tasks and models. Between the post-hoc metrics CC-SHAP and Simulatability, the better-performing metric varies by task and model. Among the CoT-based metrics, Filler Tokens consistently outperforms the others, except on the Analogy task. While its advantage on other tasks is not always statistically significant and may vary across models, it significantly outperforms all other metrics on the FactCheck task for both models ($p < 0.05$, Wilcoxon signed-rank test\footnote{Wilcoxon signed-rank test used for all statistical testing unless stated otherwise.}). To assess overall performance, we conduct pairwise comparisons across all metrics, tasks, and models using Copeland’s method. As seen in Table \ref{tab:diagnosticity}, CC-SHAP ranks highest among post-hoc metrics, while Filler Tokens leads among CoT-based metrics. Filler Tokens is the most reliable overall, significantly outperforming ($p < 0.05$, one-sample t-test) the baseline value of $0.5$ across all tasks and models. The Multi-hop task is particularly challenging, as all other metrics fail to significantly exceed baseline performance.

\subsection{Metric Sensitivity}
\label{subsec:metric_sensitivity}
\begin{figure}[t]
    \centering
    \includegraphics[width=\linewidth]{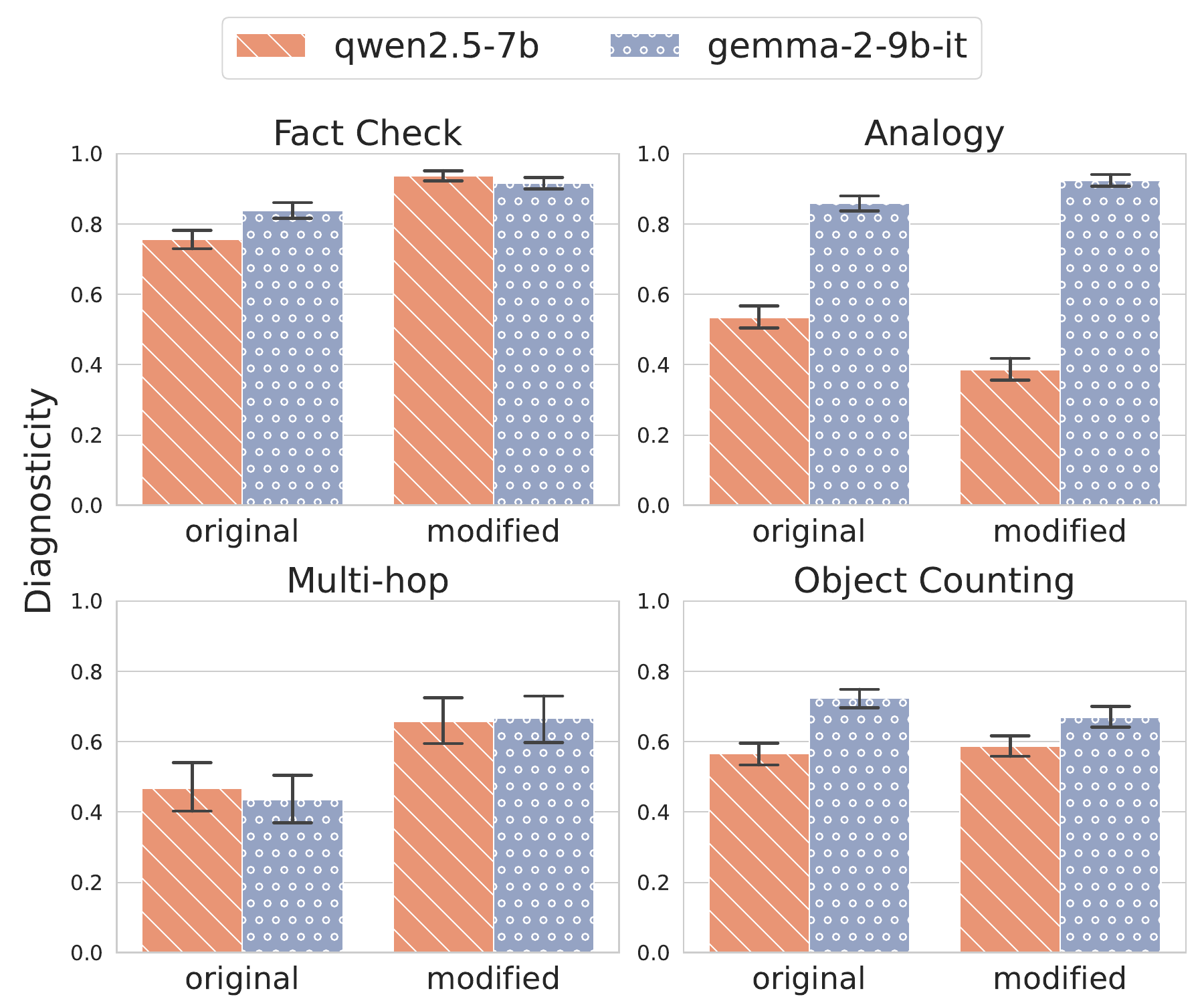}
    \caption{Comparison of original and modified Early Answering metrics across four tasks and two models: \texttt{qwen2.5-7b}, \texttt{gemma-2-9b-it}. Errorbars indicate the 95\% bootstrap confidence intervals.}
    \label{fig:early_answering_ablation}
\end{figure}

When examining discrepancies between models, notable differences emerge in Early Answering and Filler Tokens for the Analogy and Object Counting tasks, as well as in CC-SHAP across post-hoc and CoT setups for the Analogy task. These inconsistencies may stem from the way these metrics operate. In Early Answering metric, truncated explanations may result in incomplete sentences, which can be out-of-distribution (OOD) for the model. Thus, drops in prediction scores may not solely reflect the unfaithfulness but rather the model's sensitivity to OOD inputs \citep{Hooker2018ABF}. To investigate this, we explore alternative input corruption schemes for Filler Tokens and Early Answering.

\paragraph{Filler Tokens} We explore two choices: the type of filler token and the replacement strategy (repeating vs. non-repeating). The original metric replaces each character with three dots. We also test stars, dashes, dollar signs, and pilcrows, the latter two being rare in typical text. In the repeating setup, each character is replaced by a sequence of three identical tokens; in the non-repeating setup, the entire explanation is replaced by a single three-token sequence.
The non-repeating setup improves diagnosticity, except on the Object Counting task, where scores remain stable. Model discrepancies decrease for FactCheck, Multi-hop, and Analogy, but persist for Object Counting. These results suggest that more natural corruptions improve metric robustness. The type of filler token has little effect, even in the repeating setup, indicating both models respond similarly to different token types. Appendix~\ref{appendix:metrics} includes a detailed analysis. 

\paragraph{Early Answering} The original Early Answering metric retains only the first third of an explanation by character count, often producing incomplete or incoherent text. To address this, we propose a set of heuristics (detailed in Appendix~\ref{appendix:metrics}) to ensure that shortened explanations are syntactically meaningful. Figure~\ref{fig:early_answering_ablation} compares diagnosticity scores across four tasks and two models using the original and modified Early Answering metrics. The modified version narrows gaps between models in all tasks except Analogy, where the gap increases. Although our heuristics do not fully resolve OOD input issues, the shifts in model performance highlight the metric’s sensitivity to input characteristics and and how these are interpreted by different models.

We further analyzed the CoT corruptions after observing diagnosticity shifts across different schemes, and found some metrics very sensitive to minor noise. See Appendix~\ref{appendix:metrics} for details. 

\subsection{Reliability of Edits}
\label{subsec:edit_reliability}
\methodname assumes that one explanation in each pair is faithful to the evaluated model, while the other is unfaithful. While synthetic explanations in principle ensure faithfulness or unfaithfulness with respect to the edited model, their practical accuracy depends on the success of the editing method. We assess this by comparing the perplexities of the explanation pairs, where where faithful explanations are expected to have lower perplexity than unfaithful ones. 

\begin{figure}[htb]
    \centering
    \includegraphics[width=1.0\linewidth]{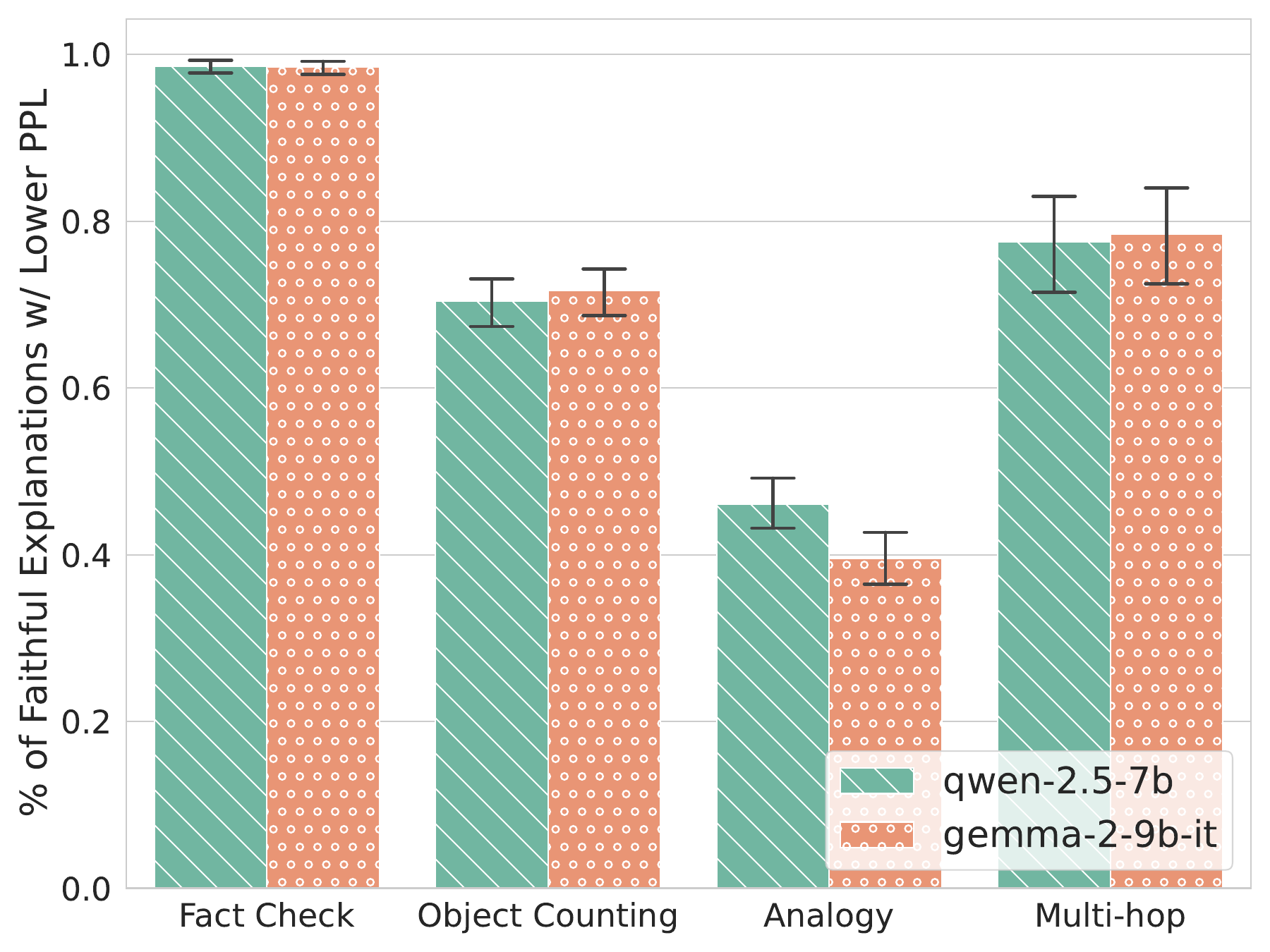}
    \caption{Percentage of faithul explanations with lower perplexity than unfaithful ones by task and model. Higher values indicates higher success in applied edits. Errorbars indicate 95\% bootstrap confidence intervals.}
    \label{fig:ppl_comparison}
\end{figure}

Figure \ref{fig:ppl_comparison} shows shows the percentage of faithful explanations with lower perplexity than unfaithful ones, by task and model. For FactCheck, the edits show strong success, with scores near 1.0, followed by Multi-hop Reasoning and Object Counting. In contrast, edits for the Analogy task underperform, with scores falling below 50\%. This is likely due to conflicting information about widely known facts, such as capital cities. To explore whether this limitation is inherent to ICE, we compare ICE and MEMIT on \texttt{qwen2.5-7b} across three tasks.  MEMIT edits show significant improvements in Analogy and Object Counting, but cause  near 50\% drop in model editing performance for FactCheck. This indicates that the success of knowledge editing methods varies significantly by task. \textit{Importantly, due to the low edit reliability scores for Analogy, ICE-based diagnosticity results for this task are not robust and should be considered unreliable.} See Appendix~\ref{appendix:results} for a detailed analysis.

\begin{table*}[t!]
        \centering
        \resizebox{\linewidth}{!}{
        \begin{tabular}{lcccccccc}
        \toprule
        \multirow{2}{*}{\textbf{Metric}} & \multicolumn{2}{c}{\textbf{FactCheck}} & \multicolumn{2}{c}{\textbf{Analogy}} & \multicolumn{2}{c}{\textbf{Object Counting}} & \multicolumn{2}{c}{\textbf{Multi-hop}} \\
        \cmidrule(lr){2-3} \cmidrule(lr){4-5} \cmidrule(lr){6-7} \cmidrule(lr){8-9}
        & \textbf{Bin.} & \textbf{Cont.} ($\Delta$)
        & \textbf{Bin.} & \textbf{Cont.} ($\Delta$)
        & \textbf{Bin.} & \textbf{Cont.} ($\Delta$)
        & \textbf{Bin.} & \textbf{Cont.} ($\Delta$) \\
        \midrule
    Early Answering & 0.496 & 0.756  \textcolor{forestgreen}{(+0.260)}& 0.501 & 0.534  \textcolor{forestgreen}{(+0.033)}& 0.488 & 0.566  \textcolor{forestgreen}{(+0.078)}& 0.488 & 0.468  \textcolor{gray}{(-0.020)}  \\
            Filler Tokens & 0.500 & 0.828  \textcolor{forestgreen}{(+0.328)}& 0.500 & 0.561  \textcolor{forestgreen}{(+0.061)}& 0.444 & 0.630  \textcolor{forestgreen}{(+0.186)}& 0.495 & 0.682  \textcolor{forestgreen}{(+0.187)}  \\
            Adding Mistakes & 0.493 & 0.534  \textcolor{forestgreen}{(+0.041)}& 0.530 & 0.590  \textcolor{forestgreen}{(+0.060)}& 0.517 & 0.614  \textcolor{forestgreen}{(+0.097)}& 0.485 & 0.542  \textcolor{gray}{(+0.057)}  \\
            Paraphrasing & 0.571 & 0.556  \textcolor{gray}{(-0.015)}& 0.501 & 0.535  \textcolor{forestgreen}{(+0.034)}& 0.531 & 0.425  \textcolor{red}{(-0.106)}& 0.510 & 0.448  \textcolor{gray}{(-0.062)}  \\
                \bottomrule
        \end{tabular}
        }
        \caption{Comparison of diagnosticity scores between continuous and binary variants of CoT corruption-based metrics using \texttt{qwen-2.5-7b}. Differences are statistically significant (Wilcoxon signed-rank test, $p < 0.05$) except those highlighted in gray.}
        \label{tab:binary_vs_continuous}
    \end{table*}

\subsection{Effect of Knowledge Editing Method}
\begin{figure}
    \centering
    \includegraphics[width=\linewidth]{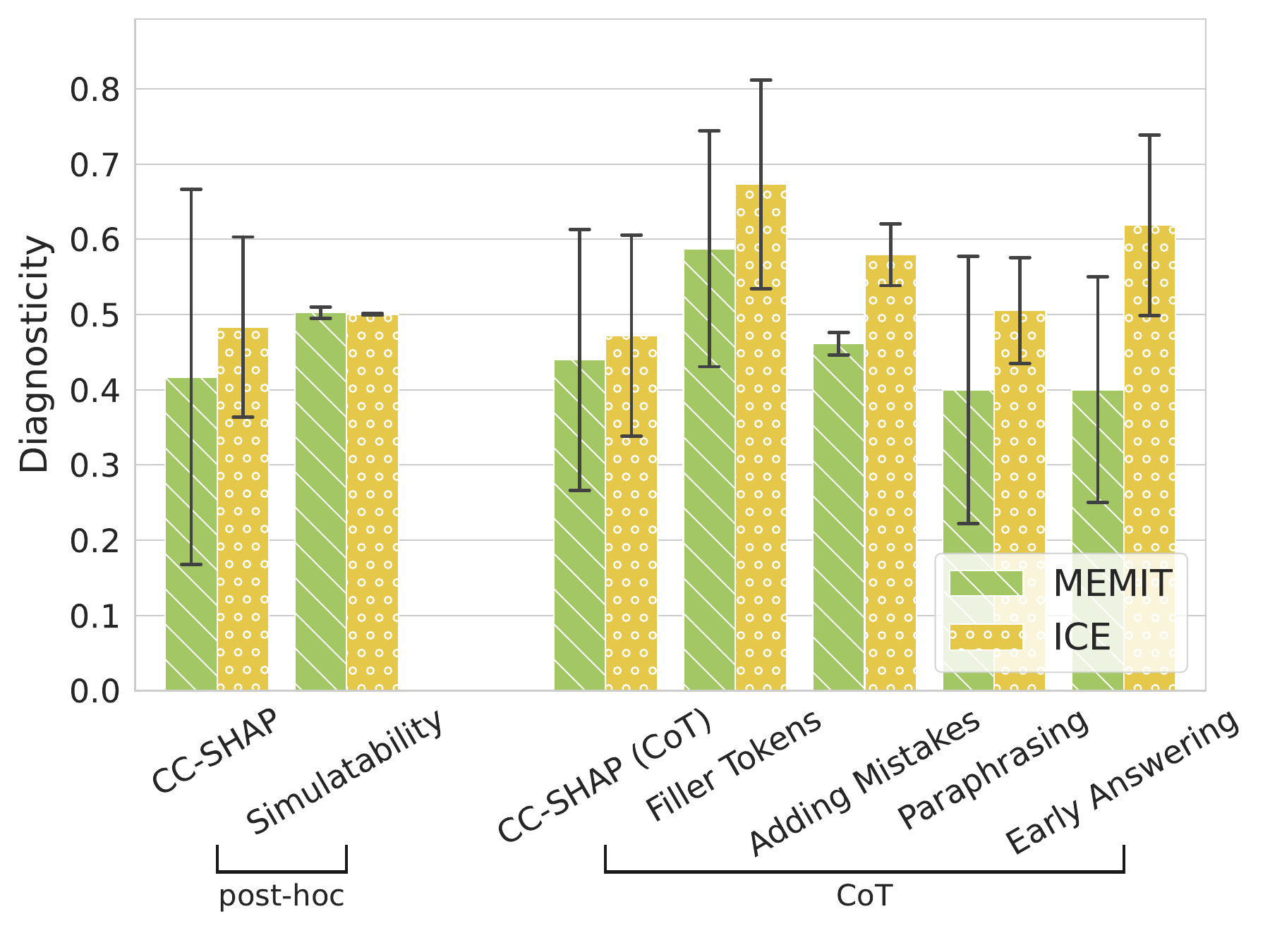}
    \caption{Diagnosticity scores for each metric on \texttt{qwen-2.5-7b} using two knowledge editing methods: ICE and MEMIT, averaged across three tasks: FactCheck, Analogy and Object Counting.}
    \label{fig:edit_methods}
\end{figure}

We replace ICE with MEMIT \citep{meng2023memit}, a locate-and-edit approach enabling bulk edits (details in Appendix \ref{appendix:model_editing}). Since Multi-hop reasoning edits do not align with MEMIT's format, this task is excluded. Figure \ref{fig:edit_methods} compares MEMIT and ICE across all faithfulness metrics, with diagnosticity scores averaged over three tasks. Except for the FactCheck task, the differences are not significant ($p > 0.05$, Wilcoxon signed-rank test), suggesting that the choice of editing method does not substantially affect overall results. Full results for MEMIT are in Appendix \ref{appendix:results}.

\subsection{Effect of Explanation Type}

\label{subsec:explanation_type}
\begin{figure}
    \centering
    \includegraphics[width=\linewidth]{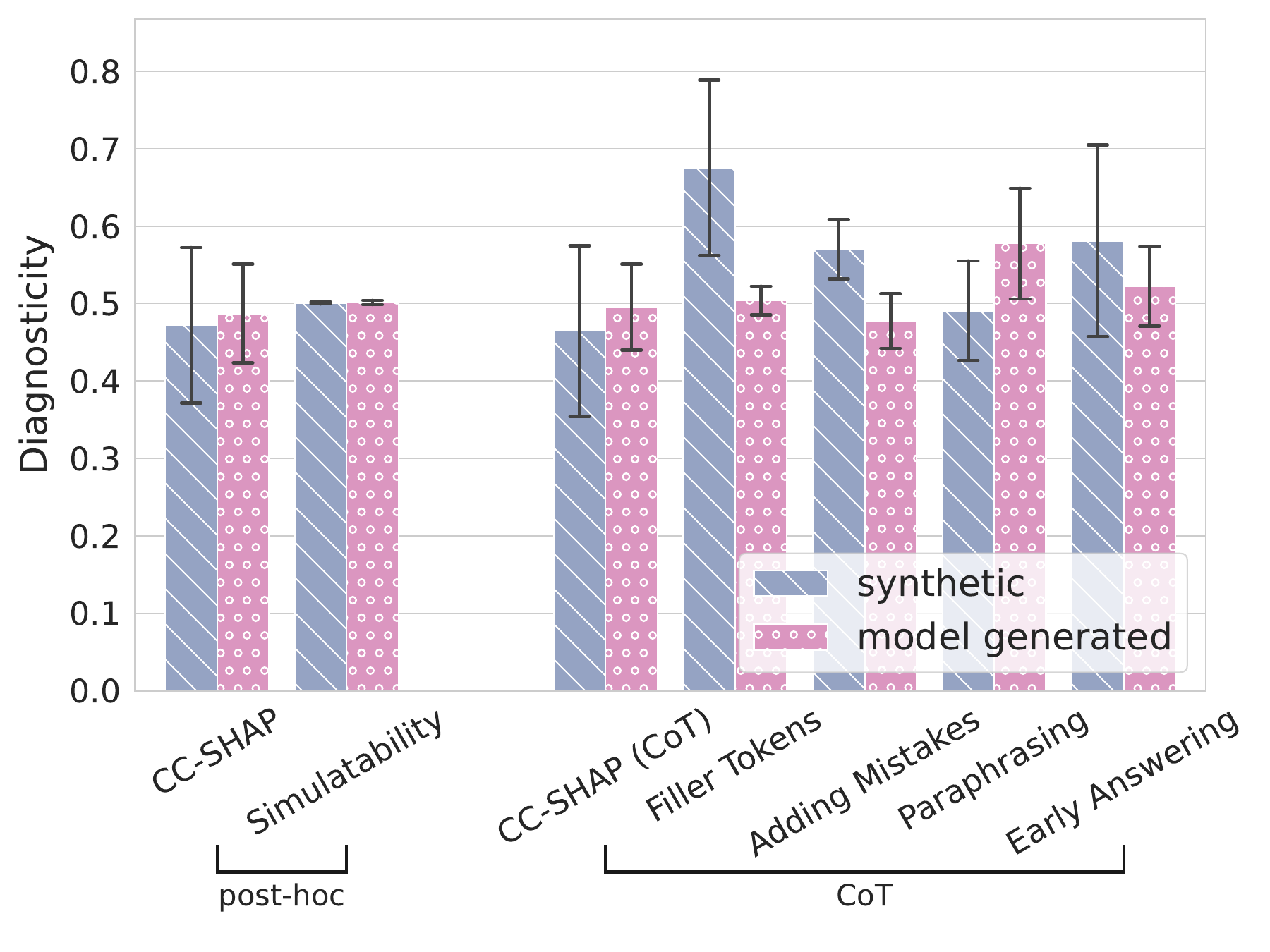}
    \caption{Diagnosticity scores for each metric using model generated and synthetic explanations with \texttt{qwen-2.5-7b}, averaged across all four tasks.
    }
    \label{fig:explanation_types}
\end{figure}

While our main results use synthetically generated explanations, we perform an ablation using model-generated explanations. We evaluate all metrics using \texttt{qwen-2.5-7b}, limiting model-generated explanations to 100 tokens. Figure \ref{fig:explanation_types} compares model-generated and synthetic explanations across faithfulness metrics, with diagnosticity scores averaged over four tasks. The results indicate that synthetic explanations generally achieve higher scores than model-generated ones, though differences across explanation types are not statistically significant ($p > 0.05$, Wilcoxon signed-rank test). Qualitatively, we find that for Analogy and Object Counting, model-generated explanations often fail to reflect the applied edits, aligning with our findings in \S \ref{subsec:edit_reliability}. Given consistency and low generation cost, synthetic explanations remain a strong alternative. 

\subsection{Binary vs. Continuous Metrics}

In Table \ref{tab:diagnosticity}, low diagnosticity scores of Simulatability, which is a metric that produces binary outcomes, are notable. 
For a more detailed analysis, we compare binary and continuous variants of CoT-based metrics across four tasks using \texttt{qwen2.5-7b}. Table \ref{tab:binary_vs_continuous} shows that continuous variants consistently outperform their binary counterparts across most tasks and metrics, with relative gains of up to 66\%. Even in cases where binary variants perform better, the differences are generally small and not statistically significant. While \citet{siegel-etal-2024-probabilities} provide theoretical rationale for preferring continuous alternatives in Counterfactual Edits \citep{Atanasova2023FaithfulnessTF}, we are the first to empirically confirm this trend across multiple metrics and tasks.

\section{Conclusion}
\label{sec:conclusion}
Our work here provides a testbed for faithfulness metrics, laying the groundwork for improvements in faithfulness metrics and natural language explanations. We benchmark popular post-hoc and CoT-based faithfulness metrics across tasks. Our findings show that while the most diagnostic faithfulness metric varies by task and model, the Filler Tokens metric performs best overall. Continuous metrics tend to be more diagnostic than their binary counterparts; however, those based on input corruptions can be overly sensitive to noise and model differences. Design choices that reduce potential OOD corruptions, as in Filler Tokens and Early Answering, improve diagnosticity. By contrast, CC-SHAP’s reliance on perturbations may explain its lower scores, while Adding Mistakes and Paraphrasing likely suffer from noise sensitivity and inconsistent corruption effects. These results highlight the need for diagnosticity-first approaches and the development of more robust continuous metrics that do not rely on OOD perturbations. 

Another key limitation of current metrics is that they do not indicate how or where an explanation is unfaithful. Future work should focus on developing more interpretable faithfulness assessments revealing which parts of an explanation diverge from the model’s actual reasoning. A recent contemporaneous work \citep{Tutek2025MeasuringCO} takes a first step in this direction by quantifying the faithfulness of reasoning steps at the sentence level. Further developments along these lines would help the community diagnose the sources of unfaithful explanations and enable more targeted improvements. Ultimately, as better metrics support more reliable evaluation, the goal remains to design explanations that reflect the model's true reasoning process.

\section*{Limitations}
\label{sec:limitations}
\methodname is not suitable for evaluating all types of faithfulness metrics. Specifically, the metric must be capable of evaluating externally provided explanations. For example, we cannot evaluate metrics like Counterfactual Edits \citep{Atanasova2023FaithfulnessTF}, which assess changes in explanations resulting from input modifications. Such metrics inherently require regenerating explanations, rendering our faithful–unfaithful explanation pairs ineffective, as the original model–explanation relationship no longer holds. Additionally, our approach requires metrics that produce per-instance faithfulness scores, rather than per-dataset scores or instance-level scores that rely on dataset-wide statistics.

Our framework also substantially depends upon the efficacy of the knowledge editing method. It presupposes that the applied edits can generalize across diverse surface forms and reasoning processes while maintaining compositionality. Previous research on knowledge editing assesses the \textit{portability} of edits by employing various benchmarks \citep{yao-etal-2023-editing, Zhong2023MQuAKEAK, cohen2024evaluating}, wherein they curate downstream applications for each specific edit. Nevertheless, the creation of such benchmarks pertinent to our tasks necessitates substantial effort, which is not within the scope of this study. Consequently, we utilize the perplexity relationship between edits and synthetically generated explanations as an indicative measure of model editing success. 

While we perform an ablation study employing MEMIT, the potential benefits of model-generated explanations and more extensive models employing alternative editing techniques remains unexamined. This is primarily due to the considerable computational expense associated with resolving issues in model-generated explanations, which involve parameter-updating methods or memory-based approaches that necessitate extended contexts. 

Additionally, our scaling experiments exclude CC-SHAP owing to its slow execution. Specifically, memory-based methods considerably extend the duration of experiments involving CC-SHAP as they increase context length.

\section*{Acknowledgements}
\label{sec:Acknowledgements}
The authors thank Peter Hase for useful pointers in knowledge editing literature, Rakesh R Menon for feedback on the paper's diagrams and All Digital, Valery Zanimanski, BomSymbols and Twemoji team for providing various icons used in Figures \ref{fig:method_overview} and \ref{fig:tasks}. This work was supported in part by NSF grant DRL2112635.

\bibliography{custom}

@inproceedings{Chan2022ACS,
  title={A Comparative Study of Faithfulness Metrics for Model Interpretability Methods},
  author={Chun Sik Chan and Huanqi Kong and Guanqing Liang},
  booktitle={Annual Meeting of the Association for Computational Linguistics},
  year={2022},
  url={https://api.semanticscholar.org/CorpusID:248118978}
}

@article{meng2023memit,
  title={Mass Editing Memory in a Transformer},
  author={Kevin Meng and Sen Sharma, Arnab and Alex Andonian and Yonatan Belinkov and David Bau},
  journal={The Eleventh International Conference on Learning Representations (ICLR)},
  year={2023}
}

@inproceedings{Jacovi2020TowardsFI,
  title={Towards Faithfully Interpretable NLP Systems: How Should We Define and Evaluate Faithfulness?},
  author={Alon Jacovi and Yoav Goldberg},
  booktitle={Annual Meeting of the Association for Computational Linguistics},
  year={2020},
  url={https://api.semanticscholar.org/CorpusID:215416110}
}

@article{Wei2022ChainOT,
  title={Chain of Thought Prompting Elicits Reasoning in Large Language Models},
  author={Jason Wei and Xuezhi Wang and Dale Schuurmans and Maarten Bosma and Ed Huai-hsin Chi and F. Xia and Quoc Le and Denny Zhou},
  journal={ArXiv},
  year={2022},
  volume={abs/2201.11903},
  url={https://api.semanticscholar.org/CorpusID:246411621}
}

@inproceedings{Parcalabescu2023OnMF,
  title={On Measuring Faithfulness or Self-consistency of Natural Language Explanations},
  author={Letitia Parcalabescu and Anette Frank},
  year={2023},
  url={https://api.semanticscholar.org/CorpusID:265150102}
}

@article{Zhang2024ACS,
  title={A Comprehensive Study of Knowledge Editing for Large Language Models},
  author={Ningyu Zhang and Yunzhi Yao and Bo Tian and Peng Wang and Shumin Deng and Meng Wang and Zekun Xi and Shengyu Mao and Jintian Zhang and Yuansheng Ni and Siyuan Cheng and Ziwen Xu and Xin Xu and Jia-Chen Gu and Yong Jiang and Pengjun Xie and Fei Huang and Lei Liang and Zhiqiang Zhang and Xiao-Jun Zhu and Jun Zhou and Huajun Chen},
  journal={ArXiv},
  year={2024},
  volume={abs/2401.01286},
  url={https://api.semanticscholar.org/CorpusID:266725300}
}

@article{Lanham2023MeasuringFI,
  title={Measuring Faithfulness in Chain-of-Thought Reasoning},
  author={Tamera Lanham and Anna Chen and Ansh Radhakrishnan and Benoit Steiner and Carson E. Denison and Danny Hernandez and Dustin Li and Esin Durmus and Evan Hubinger and John Kernion and Kamil.e Lukovsiut.e and Karina Nguyen and Newton Cheng and Nicholas Joseph and Nicholas Schiefer and Oliver Rausch and Robin Larson and Samuel McCandlish and Sandipan Kundu and Saurav Kadavath and Shannon Yang and Tom Henighan and Timothy D. Maxwell and Timothy Telleen-Lawton and Tristan Hume and Zac Hatfield-Dodds and Jared Kaplan and Janina Brauner and Sam Bowman and Ethan Perez},
  journal={ArXiv},
  year={2023},
  volume={abs/2307.13702},
  url={https://api.semanticscholar.org/CorpusID:259953372}
}

@article{Turpin2023LanguageMD,
  title={Language Models Don't Always Say What They Think: Unfaithful Explanations in Chain-of-Thought Prompting},
  author={Miles Turpin and Julian Michael and Ethan Perez and Sam Bowman},
  journal={ArXiv},
  year={2023},
  volume={abs/2305.04388},
  url={https://api.semanticscholar.org/CorpusID:258556812}
}

@article{Atanasova2023FaithfulnessTF,
  title={Faithfulness Tests for Natural Language Explanations},
  author={Pepa Atanasova and Oana-Maria Camburu and Christina Lioma and Thomas Lukasiewicz and Jakob Grue Simonsen and Isabelle Augenstein},
  journal={ArXiv},
  year={2023},
  volume={abs/2305.18029},
  url={https://api.semanticscholar.org/CorpusID:258960511}
}

@inproceedings{Wiegreffe2020MeasuringAB,
  title={Measuring Association Between Labels and Free-Text Rationales},
  author={Sarah Wiegreffe and Ana Marasovi{\'c} and Noah A. Smith},
  booktitle={Conference on Empirical Methods in Natural Language Processing},
  year={2020},
  url={https://api.semanticscholar.org/CorpusID:225068329}
}

@article{Patil2023CanSI,
  title={Can Sensitive Information Be Deleted From LLMs? Objectives for Defending Against Extraction Attacks},
  author={Vaidehi Patil and Peter Hase and Mohit Bansal},
  journal={ArXiv},
  year={2023},
  volume={abs/2309.17410},
  url={https://api.semanticscholar.org/CorpusID:263311025}
}

@inproceedings{Meng2022LocatingAE,
  title={Locating and Editing Factual Associations in GPT},
  author={Kevin Meng and David Bau and Alex Andonian and Yonatan Belinkov},
  booktitle={Neural Information Processing Systems},
  year={2022},
  url={https://api.semanticscholar.org/CorpusID:255825985}
}

@inproceedings{zheng-etal-2023-edit,
    title = "Can We Edit Factual Knowledge by In-Context Learning?",
    author = "Zheng, Ce  and
      Li, Lei  and
      Dong, Qingxiu  and
      Fan, Yuxuan  and
      Wu, Zhiyong  and
      Xu, Jingjing  and
      Chang, Baobao",
    editor = "Bouamor, Houda  and
      Pino, Juan  and
      Bali, Kalika",
    booktitle = "Proceedings of the 2023 Conference on Empirical Methods in Natural Language Processing",
    month = dec,
    year = "2023",
    address = "Singapore",
    publisher = "Association for Computational Linguistics",
    url = "https://aclanthology.org/2023.emnlp-main.296",
    doi = "10.18653/v1/2023.emnlp-main.296",
    pages = "4862--4876",
}

@inproceedings{gupta-etal-2023-editing,
    title = "Editing Common Sense in Transformers",
    author = "Gupta, Anshita  and
      Mondal, Debanjan  and
      Sheshadri, Akshay  and
      Zhao, Wenlong  and
      Li, Xiang  and
      Wiegreffe, Sarah  and
      Tandon, Niket",
    editor = "Bouamor, Houda  and
      Pino, Juan  and
      Bali, Kalika",
    booktitle = "Proceedings of the 2023 Conference on Empirical Methods in Natural Language Processing",
    month = dec,
    year = "2023",
    address = "Singapore",
    publisher = "Association for Computational Linguistics",
    url = "https://aclanthology.org/2023.emnlp-main.511",
    doi = "10.18653/v1/2023.emnlp-main.511",
    pages = "8214--8232",
}

@inproceedings{mitchell2022fast,
    title={Fast Model Editing at Scale},
    author={Eric Mitchell and Charles Lin and Antoine Bosselut and Chelsea Finn and Christopher D Manning},
    booktitle={International Conference on Learning Representations},
    year={2022},
    url={https://openreview.net/pdf?id=0DcZxeWfOPt}
}

@inproceedings{NEURIPS2023_3927bbdc,
 author = {Hase, Peter and Bansal, Mohit and Kim, Been and Ghandeharioun, Asma},
 booktitle = {Advances in Neural Information Processing Systems},
 editor = {A. Oh and T. Naumann and A. Globerson and K. Saenko and M. Hardt and S. Levine},
 pages = {17643--17668},
 publisher = {Curran Associates, Inc.},
 title = {Does Localization Inform Editing? Surprising Differences in Causality-Based Localization vs. Knowledge Editing in Language Models},
 url = {https://proceedings.neurips.cc/paper_files/paper/2023/file/3927bbdcf0e8d1fa8aa23c26f358a281-Paper-Conference.pdf},
 volume = {36},
 year = {2023}
}

@inproceedings{geva-etal-2023-dissecting,
    title = "Dissecting Recall of Factual Associations in Auto-Regressive Language Models",
    author = "Geva, Mor  and
      Bastings, Jasmijn  and
      Filippova, Katja  and
      Globerson, Amir",
    editor = "Bouamor, Houda  and
      Pino, Juan  and
      Bali, Kalika",
    booktitle = "Proceedings of the 2023 Conference on Empirical Methods in Natural Language Processing",
    month = dec,
    year = "2023",
    address = "Singapore",
    publisher = "Association for Computational Linguistics",
    url = "https://aclanthology.org/2023.emnlp-main.751",
    doi = "10.18653/v1/2023.emnlp-main.751",
    pages = "12216--12235",
}

@article{cohen2024evaluating,
  title={Evaluating the ripple effects of knowledge editing in language models},
  author={Cohen, Roi and Biran, Eden and Yoran, Ori and Globerson, Amir and Geva, Mor},
  journal={Transactions of the Association for Computational Linguistics},
  volume={12},
  pages={283--298},
  year={2024},
  publisher={MIT Press One Broadway, 12th Floor, Cambridge, Massachusetts 02142, USA~…}
}

@inproceedings{hartvigsen2023aging,
  title={Aging with GRACE: Lifelong Model Editing with Discrete Key-Value Adaptors},
  author={Hartvigsen, Thomas and Sankaranarayanan, Swami and Palangi, Hamid and Kim, Yoon and Ghassemi, Marzyeh},
  booktitle={Advances in Neural Information Processing Systems},
  year={2023}
}

@article{srivastava2023beyond,
  title={Beyond the Imitation Game: Quantifying and extrapolating the capabilities of language models},
  author={BIG\-{B}ench authors},
  journal={Transactions on Machine Learning Research},
  issn={2835-8856},
  year={2023},
  url={https://openreview.net/forum?id=uyTL5Bvosj},
  note={}
}

@inproceedings{Lundberg2017AUA,
  title={A Unified Approach to Interpreting Model Predictions},
  author={Scott M. Lundberg and Su-In Lee},
  booktitle={Neural Information Processing Systems},
  year={2017},
  url={https://api.semanticscholar.org/CorpusID:21889700}
}

@inproceedings{Hase2020LeakageAdjustedSC,
  title={Leakage-Adjusted Simulatability: Can Models Generate Non-Trivial Explanations of Their Behavior in Natural Language?},
  author={Peter Hase and Shiyue Zhang and Harry Xie and Mohit Bansal},
  booktitle={Findings},
  year={2020},
  url={https://api.semanticscholar.org/CorpusID:222209056}
}

@inproceedings{Hase2020EvaluatingEA,
  title={Evaluating Explainable AI: Which Algorithmic Explanations Help Users Predict Model Behavior?},
  author={Peter Hase and Mohit Bansal},
  booktitle={Annual Meeting of the Association for Computational Linguistics},
  year={2020},
  url={https://api.semanticscholar.org/CorpusID:218502350}
}

@article{DoshiVelez2017TowardsAR,
  title={Towards A Rigorous Science of Interpretable Machine Learning},
  author={Finale Doshi-Velez and Been Kim},
  journal={arXiv: Machine Learning},
  year={2017},
  url={https://api.semanticscholar.org/CorpusID:11319376}
}

@inproceedings{Chan2022FRAMEER,
  title={FRAME: Evaluating Rationale-Label Consistency Metrics for Free-Text Rationales},
  author={Aaron Chan and Shaoliang Nie and Liang Tan and Xiaochang Peng and Hamed Firooz and Maziar Sanjabi and Xiang Ren},
  year={2022},
  url={https://api.semanticscholar.org/CorpusID:254247321}
}

@inproceedings{yao-etal-2023-editing,
    title = "Editing Large Language Models: Problems, Methods, and Opportunities",
    author = "Yao, Yunzhi  and
      Wang, Peng  and
      Tian, Bozhong  and
      Cheng, Siyuan  and
      Li, Zhoubo  and
      Deng, Shumin  and
      Chen, Huajun  and
      Zhang, Ningyu",
    editor = "Bouamor, Houda  and
      Pino, Juan  and
      Bali, Kalika",
    booktitle = "Proceedings of the 2023 Conference on Empirical Methods in Natural Language Processing",
    month = dec,
    year = "2023",
    address = "Singapore",
    publisher = "Association for Computational Linguistics",
    url = "https://aclanthology.org/2023.emnlp-main.632",
    doi = "10.18653/v1/2023.emnlp-main.632",
    pages = "10222--10240",
}

@article{Geva2021DidAU,
  title={Did Aristotle Use a Laptop? A Question Answering Benchmark with Implicit Reasoning Strategies},
  author={Mor Geva and Daniel Khashabi and Elad Segal and Tushar Khot and Dan Roth and Jonathan Berant},
  journal={Transactions of the Association for Computational Linguistics},
  year={2021},
  volume={9},
  pages={346-361},
  url={https://api.semanticscholar.org/CorpusID:230799347}
}

@article{Yang2024Qwen25TR,
  title={Qwen2.5 Technical Report},
  author={Qwen An Yang and Baosong Yang and Beichen Zhang and Binyuan Hui and Bo Zheng and Bowen Yu and Chengyuan Li and Dayiheng Liu and Fei Huang and Haoran Wei and Huan Lin and Jian Yang and Jianhong Tu and Jianwei Zhang and Jianxin Yang and Jiaxin Yang and Jingren Zhou and Junyang Lin and Kai Dang and Keming Lu and Keqin Bao and Kexin Yang and Le Yu and Mei Li and Mingfeng Xue and Pei Zhang and Qin Zhu and Rui Men and Runji Lin and Tianhao Li and Tingyu Xia and Xingzhang Ren and Xuancheng Ren and Yang Fan and Yang Su and Yi-Chao Zhang and Yunyang Wan and Yuqi Liu and Zeyu Cui and Zhenru Zhang and Zihan Qiu and Shanghaoran Quan},
  journal={ArXiv},
  year={2024},
  volume={abs/2412.15115},
  url={https://api.semanticscholar.org/CorpusID:274859421}
}

@article{Riviere2024Gemma2I,
  title={Gemma 2: Improving Open Language Models at a Practical Size},
  author={Gemma Team Morgane Riviere and Shreya Pathak and Pier Giuseppe Sessa and Cassidy Hardin and Surya Bhupatiraju and L'eonard Hussenot and Thomas Mesnard and Bobak Shahriari and Alexandre Ram'e and Johan Ferret and Peter Liu and Pouya Dehghani Tafti and Abe Friesen and Michelle Casbon and Sabela Ramos and Ravin Kumar and Charline Le Lan and Sammy Jerome and Anton Tsitsulin and Nino Vieillard and Piotr Stańczyk and Sertan Girgin and Nikola Momchev and Matt Hoffman and Shantanu Thakoor and Jean-Bastien Grill and Behnam Neyshabur and Alanna Walton and Aliaksei Severyn and Alicia Parrish and Aliya Ahmad and Allen Hutchison and Alvin Abdagic and Amanda Carl and Amy Shen and Andy Brock and Andy Coenen and Anthony Laforge and Antonia Paterson and Ben Bastian and Bilal Piot and Boxi Wu and Brandon Royal and Charlie Chen and Chintu Kumar and Chris Perry and Christoper A. Welty and Christopher A. Choquette-Choo and Danila Sinopalnikov and David Weinberger and Dimple Vijaykumar and Dominika Rogozi'nska and D. Herbison and Elisa Bandy and Emma Wang and Eric Noland and Erica Moreira and Evan Senter and Evgenii Eltyshev and Francesco Visin and Gabriel Rasskin and Gary Wei and Glenn Cameron and Gus Martins and Hadi Hashemi and Hanna Klimczak-Pluci'nska and Harleen Batra and Harsh Dhand and Ivan Nardini and Jacinda Mein and Jack Zhou and James Svensson and Jeff Stanway and Jetha Chan and Jin Zhou and Joana Carrasqueira and Joana Iljazi and Jocelyn Becker and Joe Fernandez and Joost R. van Amersfoort and Josh Gordon and Josh Lipschultz and Joshua Newlan and Junsong Ji and Kareem Mohamed and Kartikeya Badola and Kat Black and Katie Millican and Keelin McDonell and Kelvin Nguyen and Kiranbir Sodhia and Kish Greene and Lars Lowe Sjoesund and Lauren Usui and L. Sifre and L. Heuermann and Leticia Lago and Lilly McNealus and Livio Baldini Soares and Logan Kilpatrick and Lucas Dixon and Luciano Martins and Machel Reid and Manvinder Singh and Mark Iverson and Martin Gorner and Mat Velloso and Mateo Wirth and Matt Davidow and Matt Miller and Matthew Rahtz and Matthew Watson and Meg Risdal and Mehran Kazemi and Michael Moynihan and Ming Zhang and Minsuk Kahng and Minwoo Park and Mofi Rahman and Mohit Khatwani and Natalie Dao and Nenshad Bardoliwalla and Nesh Devanathan and Neta Dumai and Nilay Chauhan and Oscar Wahltinez and Pankil Botarda and Parker Barnes and Paul Barham and Paul Michel and Pengchong Jin and Petko Georgiev and Phil Culliton and Pradeep Kuppala and Ramona Comanescu and Ramona Merhej and Reena Jana and Reza Rokni and Rishabh Agarwal and Ryan Mullins and Samaneh Saadat and S. Mc Carthy and Sarah Perrin and S'ebastien M. R. Arnold and Se-bastian Krause and Shengyang Dai and Shruti Garg and Shruti Sheth and Sue Ronstrom and Susan Chan and Timothy Jordan and Ting Yu and Tom Eccles and Tom Hennigan and Tom{\'a}s Kocisk{\'y} and Tulsee Doshi and Vihan Jain and Vikas Yadav and Vilobh Meshram and Vishal Dharmadhikari and Warren Barkley and Wei Wei and Wenming Ye and Woohyun Han and Woosuk Kwon and Xiang Xu and Zhe Shen and Zhitao Gong and Zichuan Wei and Victor Cotruta and Phoebe Kirk and Anand Rao and Minh Giang and Ludovic Peran and Tris Warkentin and Eli Collins and Joelle Barral and Zoubin Ghahramani and Raia Hadsell and D. Sculley and Jeanine Banks and Anca Dragan and Slav Petrov and Oriol Vinyals and Jeffrey Dean and Demis Hassabis and Koray Kavukcuoglu and Cl{\'e}ment Farabet and Elena Buchatskaya and Sebastian Borgeaud and Noah Fiedel and Armand Joulin and Kathleen Kenealy and Robert Dadashi and Alek Andreev},
  journal={ArXiv},
  year={2024},
  volume={abs/2408.00118},
  url={https://api.semanticscholar.org/CorpusID:270843326}
}

@article{Cohen2023EvaluatingTR,
  title={Evaluating the Ripple Effects of Knowledge Editing in Language Models},
  author={Roi Cohen and Eden Biran and Ori Yoran and Amir Globerson and Mor Geva},
  journal={Transactions of the Association for Computational Linguistics},
  year={2023},
  volume={12},
  pages={283-298},
  url={https://api.semanticscholar.org/CorpusID:260356612}
}

@article{Wang2023EasyEditAE,
  title={EasyEdit: An Easy-to-use Knowledge Editing Framework for Large Language Models},
  author={Peng Wang and Ningyu Zhang and Xin Xie and Yunzhi Yao and Bo Tian and Mengru Wang and Zekun Xi and Siyuan Cheng and Kangwei Liu and Yuansheng Ni and Guozhou Zheng and Huajun Chen},
  journal={ArXiv},
  year={2023},
  volume={abs/2308.07269},
  url={https://api.semanticscholar.org/CorpusID:260887090}
}

@inproceedings{siegel-etal-2024-probabilities,
    title = "The Probabilities Also Matter: A More Faithful Metric for Faithfulness of Free-Text Explanations in Large Language Models",
    author = "Siegel, Noah  and
      Camburu, Oana-Maria  and
      Heess, Nicolas  and
      Perez-Ortiz, Maria",
    editor = "Ku, Lun-Wei  and
      Martins, Andre  and
      Srikumar, Vivek",
    booktitle = "Proceedings of the 62nd Annual Meeting of the Association for Computational Linguistics (Volume 2: Short Papers)",
    month = aug,
    year = "2024",
    address = "Bangkok, Thailand",
    publisher = "Association for Computational Linguistics",
    url = "https://aclanthology.org/2024.acl-short.49/",
    doi = "10.18653/v1/2024.acl-short.49",
    pages = "530--546"
}

@article{Zhu2020ModifyingMI,
  title={Modifying Memories in Transformer Models},
  author={Chen Zhu and Ankit Singh Rawat and Manzil Zaheer and Srinadh Bhojanapalli and Daliang Li and Felix X. Yu and Sanjiv Kumar},
  journal={ArXiv},
  year={2020},
  volume={abs/2012.00363},
  url={https://api.semanticscholar.org/CorpusID:227238659}
}

@inproceedings{DeCao2021EditingFK,
  title={Editing Factual Knowledge in Language Models},
  author={Nicola De Cao and Wilker Aziz and Ivan Titov},
  booktitle={Conference on Empirical Methods in Natural Language Processing},
  year={2021},
  url={https://api.semanticscholar.org/CorpusID:233289412}
}

@article{Pearl2001DirectAI,
  title={Direct and Indirect Effects},
  author={Judea Pearl},
  journal={Probabilistic and Causal Inference},
  year={2001},
  url={https://api.semanticscholar.org/CorpusID:5947965}
}

@inproceedings{Vig2020InvestigatingGB,
  title={Investigating Gender Bias in Language Models Using Causal Mediation Analysis},
  author={Jesse Vig and Sebastian Gehrmann and Yonatan Belinkov and Sharon Qian and Daniel Nevo and Yaron Singer and Stuart M. Shieber},
  booktitle={Neural Information Processing Systems},
  year={2020},
  url={https://api.semanticscholar.org/CorpusID:227275068}
}

@inproceedings{Zhong2023MQuAKEAK,
  title={MQuAKE: Assessing Knowledge Editing in Language Models via Multi-Hop Questions},
  author={Zexuan Zhong and Zhengxuan Wu and Christopher D. Manning and Christopher Potts and Danqi Chen},
  booktitle={Conference on Empirical Methods in Natural Language Processing},
  year={2023},
  url={https://api.semanticscholar.org/CorpusID:258865984}
}

@inproceedings{Gu2023PokeMQAPK,
  title={PokeMQA: Programmable knowledge editing for Multi-hop Question Answering},
  author={Hengrui Gu and Kaixiong Zhou and Xiaotian Han and Ninghao Liu and Ruobing Wang and Xin Wang},
  booktitle={Annual Meeting of the Association for Computational Linguistics},
  year={2023},
  url={https://api.semanticscholar.org/CorpusID:266551175}
}

@inproceedings{mitchell2022memory,
    title={Memory-Based Model Editing at Scale},
    author={Mitchell, Eric and Lin, Charles and Bosselut, Antoine and Finn, Chelsea and Manning, Christopher D.},
    booktitle={International Conference on Machine Learning},
    url={https://arxiv.org/pdf/2206.06520.pdf},
    year={2022},
}

@inproceedings{wang-etal-2024-easyedit,
    title = "{E}asy{E}dit: An Easy-to-use Knowledge Editing Framework for Large Language Models",
    author = "Wang, Peng  and
      Zhang, Ningyu  and
      Tian, Bozhong  and
      Xi, Zekun  and
      Yao, Yunzhi  and
      Xu, Ziwen  and
      Wang, Mengru  and
      Mao, Shengyu  and
      Wang, Xiaohan  and
      Cheng, Siyuan  and
      Liu, Kangwei  and
      Ni, Yuansheng  and
      Zheng, Guozhou  and
      Chen, Huajun",
    editor = "Cao, Yixin  and
      Feng, Yang  and
      Xiong, Deyi",
    booktitle = "Proceedings of the 62nd Annual Meeting of the Association for Computational Linguistics (Volume 3: System Demonstrations)",
    month = aug,
    year = "2024",
    address = "Bangkok, Thailand",
    publisher = "Association for Computational Linguistics",
    url = "https://aclanthology.org/2024.acl-demos.9/",
    doi = "10.18653/v1/2024.acl-demos.9",
    pages = "82--93"
}

@inproceedings{lin2023awq,
 author = {Lin, Ji and Tang, Jiaming and Tang, Haotian and Yang, Shang and Chen, Wei-Ming and Wang, Wei-Chen and Xiao, Guangxuan and Dang, Xingyu and Gan, Chuang and Han, Song},
 booktitle = {Proceedings of Machine Learning and Systems},
 editor = {P. Gibbons and G. Pekhimenko and C. De Sa},
 pages = {87--100},
 title = {{AWQ}: Activation-aware Weight Quantization for On-Device LLM Compression and Acceleration},
 url = {https://proceedings.mlsys.org/paper_files/paper/2024/file/42a452cbafa9dd64e9ba4aa95cc1ef21-Paper-Conference.pdf},
 volume = {6},
 year = {2024}
}

@inproceedings{Hooker2018ABF,
  title={A Benchmark for Interpretability Methods in Deep Neural Networks},
  author={Sara Hooker and D. Erhan and Pieter-Jan Kindermans and Been Kim},
  booktitle={Neural Information Processing Systems},
  year={2018},
  url={https://api.semanticscholar.org/CorpusID:202782699}
}

@inproceedings{Tutek2025MeasuringCO,
  title={Measuring Chain of Thought Faithfulness by Unlearning Reasoning Steps},
  author={Martin Tutek and Fateme Hashemi Chaleshtori and Ana Marasovi'c and Yonatan Belinkov},
  year={2025},
  url={https://api.semanticscholar.org/CorpusID:276482396}
}

@article{Lyu2022TowardsFM,
  title={Towards Faithful Model Explanation in NLP: A Survey},
  author={Qing Lyu and Marianna Apidianaki and Chris Callison-Burch},
  journal={Computational Linguistics},
  year={2022},
  volume={50},
  pages={657-723},
  url={https://api.semanticscholar.org/CorpusID:252519203}
}

\appendix
\label{sec:appendix}
\section{Effect of Model Size}
\label{appendix:model_size}
\begin{figure}
    \centering
    \includegraphics[width=\linewidth]{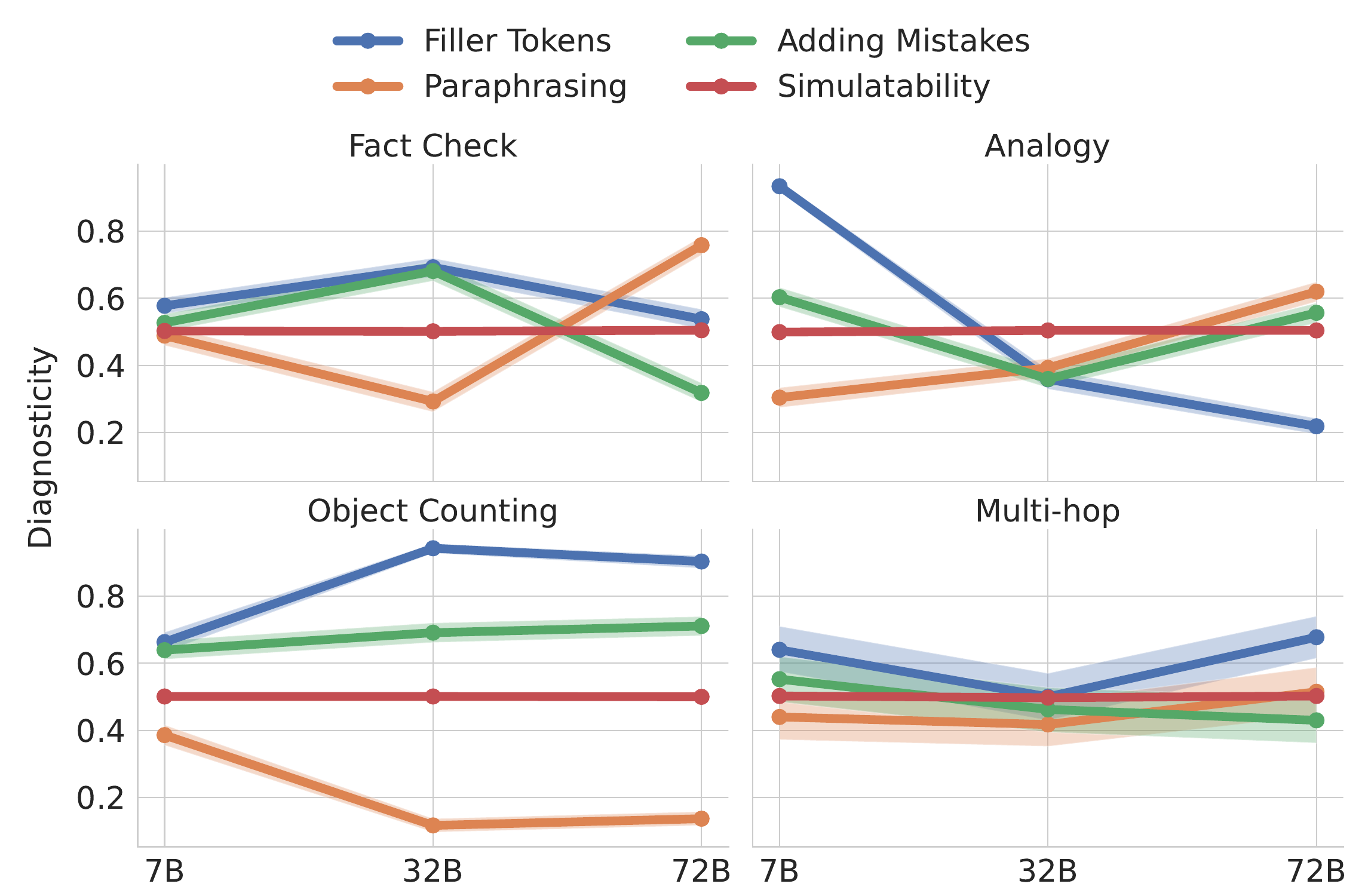}
    \caption{Comparison of diagnosticity scores with respect to model size for four metrics using 7B, 32B and 72B \texttt{qwen2.5-instruct} models. Shaded regions indicate the 95\% confidence interval calculated by bootstrap.}
    \label{fig:size_effect}
\end{figure}

\begin{table*}[t!]
        \centering
        \resizebox{0.9\linewidth}{!}{
        \begin{tabular}{lcccc}
        \toprule
        \textbf{Metric} & \textbf{FactCheck} & \textbf{Analogy} & \textbf{Object Counting} & \textbf{Multi-Hop} \\
        \midrule
    \multicolumn{5}{c}{\textbf{7B}} \\ 
 \midrule 
    Simulatability & 0.502 & 0.499 & 0.501 & 0.502   \\
                    Filler Tokens & \underline{0.578} & \underline{0.934} & \underline{0.663} & \underline{0.640}   \\
                    Adding Mistakes & \underline{0.526} & \underline{0.603} & \underline{0.639} & 0.552   \\
                    Paraphrasing & 0.488 & 0.304 & 0.386 & 0.440   \\
                
 \midrule 
\multicolumn{5}{c}{\textbf{32B}} \\ 
 \midrule 
    Simulatability & 0.501 & \underline{0.504}  \textcolor{forestgreen}{(+0.01)}& 0.501 & 0.498   \\
                    Filler Tokens & \underline{0.692}  \textcolor{forestgreen}{(+0.11)}& 0.358  \textcolor{red}{(--0.58)}& \underline{0.942}  \textcolor{forestgreen}{(+0.28)}& 0.500  \textcolor{red}{(--0.14)}  \\
                    Adding Mistakes & \underline{0.681}  \textcolor{forestgreen}{(+0.16)}& 0.360  \textcolor{red}{(--0.24)}& \underline{0.691}  \textcolor{forestgreen}{(+0.05)}& 0.462  \textcolor{red}{(--0.09)}  \\
                    Paraphrasing & 0.292  \textcolor{red}{(--0.20)}& 0.392  \textcolor{forestgreen}{(+0.09)}& 0.117  \textcolor{red}{(--0.27)}& 0.418  \textcolor{red}{(--0.02)}  \\
                
 \midrule 
\multicolumn{5}{c}{\textbf{72B}} \\ 
 \midrule  
    Simulatability & 0.504 & \underline{0.504}  \textcolor{forestgreen}{(+0.01)}& 0.500 & 0.502   \\
                    Filler Tokens & \underline{0.538}  \textcolor{red}{(--0.04)}& 0.218  \textcolor{red}{(--0.72)}& \underline{0.903}  \textcolor{forestgreen}{(+0.24)}& \underline{0.678}  \textcolor{forestgreen}{(+0.04)}  \\
                    Adding Mistakes & 0.318  \textcolor{red}{(--0.21)}& \underline{0.556}  \textcolor{red}{(--0.05)}& \underline{0.711}  \textcolor{forestgreen}{(+0.07)}& 0.430  \textcolor{red}{(--0.12)}  \\
                    Paraphrasing & \underline{0.758}  \textcolor{forestgreen}{(+0.27)}& \underline{0.620}  \textcolor{forestgreen}{(+0.32)}& 0.137  \textcolor{red}{(--0.25)}& 0.515  \textcolor{forestgreen}{(+0.08)}  \\
                
    \bottomrule
        \end{tabular}
        }
        \caption{The change in diagnosticity scores across with respect to model size across four tasks. Underlined numbers show the diagnosticity scores that are significantly higher than $0.5$ (one-sample t-test, $p < 0.05$).}
        \label{tab:scaling_diagnosticity}
    \end{table*}

Our main experiments are conducted on relatively small models with 7 billion to 9 billion parameters. We evaluate the impact of model size on diagnosticity by testing Simulatability, Filler Tokens, Adding Mistakes, and Paraphrasing on three models: \texttt{qwen2.5-7b-instruct}, \texttt{qwen2.5-32b-instruct}, and \texttt{qwen2.5-72b-instruct}. For the 32B and 72B models we adopt their AWQ \citep{lin2023awq} versions due to memory considerations. Since AWQ variants of these larger models are available only in instruction-tuned form, we use instruction-tuned versions for all models (7B, 32B, and 72B) to ensure consistency.

\begin{figure}[htb]
    \centering
    \includegraphics[width=\linewidth]{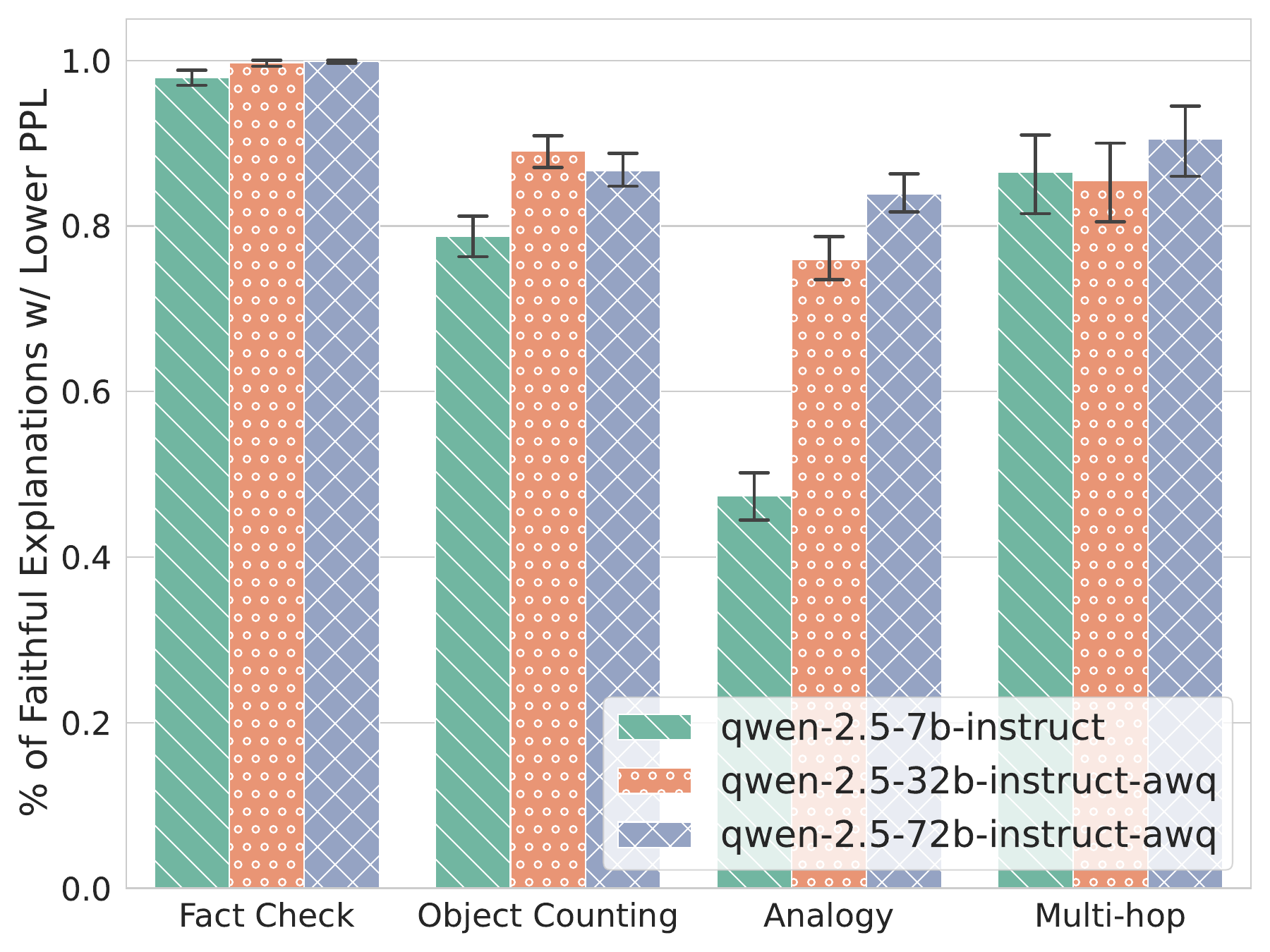}
    \caption{Comparison of the edit reliability across four tasks using models of varying sizes: \texttt{qwen2.5-7b-instruct}, \texttt{qwen2.5-32b-instruct-awq}, \texttt{qwen2.5-72b-instruct-awq}. A higher frequency indicates greater success in applied edits. Errorbars indicate the 95\% confidence interval calculated by bootstrap.}
    \label{fig:ppl_comparison_size}
\end{figure}

Figure \ref{fig:size_effect} shows no clear scaling trends in diagnosticity. Simulatability remains stable, while Adding Mistakes shows slight improvements with scale in the Object Counting task but mixed patterns for other tasks. Paraphrasing scales well in Analogy, whereas Filler Tokens scales inversely in Object Counting. While Figure \ref{fig:ppl_comparison_size}  suggests edit reliability improves with model size, our results indicate that scaling shows no uniform patterns across different configurations.

Table \ref{tab:scaling_diagnosticity} examines how diagnosticity scores vary with model size for selected metrics.

\section{Additional Results}
\label{appendix:results}

\begin{figure}[htb]
    \centering
    \includegraphics[width=\linewidth]{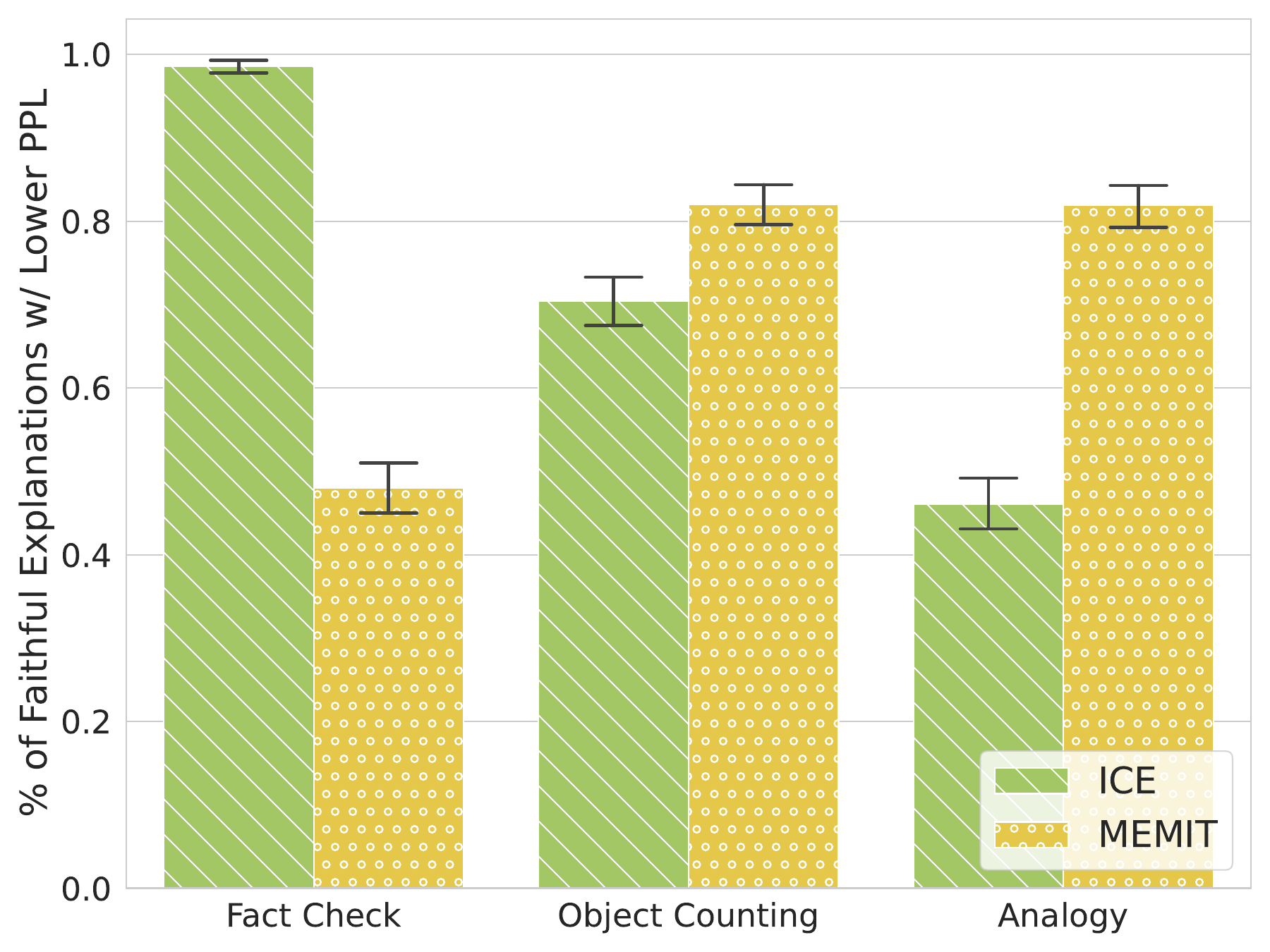}
    \caption{Comparison of the edit reliability of two editing methods across three tasks using \texttt{qwen2.5-7b}. A higher frequency indicates greater success in applied edits. Errorbars indicate the 95\% confidence interval calculated by bootstrap.}
    \label{fig:ppl_comparison_memit_vs_ice}
\end{figure}

Table \ref{tab:binary_vs_continuous} compares the binary and continuous variants of CoT-corruption-based metrics. Table \ref{tab:diagnosticity_memit} reports the diagnosticity scores when the knowledge editing method is switched from ICE to MEMIT, while Table \ref{tab:diagnosticity_model_generated} presents the scores when using model-generated explanations instead of synthetic ones. Additionally, Figure \ref{fig:ppl_comparison_memit_vs_ice} compares MEMIT and ICE in terms of edit success across three tasks.

\begin{table*}[t!]
    \tiny
        \centering
        \resizebox{0.9\linewidth}{!}{
        \begin{tabular}{lcccccc}
        \toprule
        \textbf{Metric} & \textbf{FactCheck} & \textbf{Analogy} & \textbf{Object Counting} & {\textbf{\makecell{Copeland \\ Score ($\uparrow$)}}} \\
        \midrule

        \multicolumn{5}{l}{\textbf{post-hoc}} \\
        \quad CC-SHAP & \underline{0.541} & 0.130 & \underline{0.580}  & 2  \\
                \quad Simulatability & 0.496 & \underline{0.511} & 0.500  & 1  \\
                \midrule

        \multicolumn{5}{l}{\textbf{CoT}} \\
        \quad Early Answering & 0.485 & 0.227 & 0.488  & 3  \\
                \quad Filler Tokens & 0.498 & \underline{0.768} & 0.496  & 9.5  \\
                \quad Adding Mistakes & 0.476 & 0.460 & 0.447  & 3  \\
                \quad Paraphrasing & 0.498 & 0.194 & 0.507  & 6.5  \\
                \quad CC-SHAP & 0.493 & 0.246 & \underline{0.580}  & 8  \\
                \bottomrule
        \end{tabular}
        }
        \caption{The diagnosticity scores of each metric across three tasks using \texttt{qwen2.5-7b} as model and \textbf{MEMIT as knowledge editing method}. Bold numbers indicate the highest scores on each task across the two categories of faithfulness metrics: post-hoc and CoT. Underlined numbers show the diagnosticity scores that are significantly higher than $0.5$ (one-sample t-test, $p < 0.05$).}
        \label{tab:diagnosticity_memit}
    \end{table*}

\begin{table*}[t!]
        \centering
        \resizebox{0.9\linewidth}{!}{
        \begin{tabular}{lcccccc}
        \toprule
        \textbf{Metric} & \textbf{FactCheck} & \textbf{Analogy} & \textbf{Object Counting} & \textbf{Multi-Hop} & {\textbf{\makecell{Copeland \\ Score ($\uparrow$)}}} \\
        \midrule

        \multicolumn{5}{l}{\textbf{post-hoc}} \\
        \quad CC-SHAP & \underline{0.562} & 0.516 & 0.451 & 0.420  & 2  \\
                \quad Simulatability & 0.505 & 0.500 & 0.500 & 0.500  & 2  \\
                \midrule

        \multicolumn{5}{l}{\textbf{CoT}} \\
        \quad Early Answering & 0.505 & \underline{0.598} & 0.501 & 0.485  & 9  \\
                \quad Filler Tokens & 0.485 & 0.495 & 0.528 & 0.507  & 7  \\
                \quad Adding Mistakes & 0.476 & 0.514 & 0.489 & 0.430  & 4  \\
                \quad Paraphrasing & \underline{0.596} & 0.510 & \underline{0.534} & \underline{0.670}  & 14  \\
                \quad CC-SHAP & 0.510 & 0.452 & 0.452 & 0.568  & 6  \\
                \bottomrule
        \end{tabular}
        }
        \caption{Diagnosticity scores of each metric across three tasks using \texttt{qwen2.5-7b} as the model and ICE as the knowledge editing method, \textbf{with model-generated explanations}. Bold numbers indicate the highest scores on each task across the two categories of faithfulness
metrics: post-hoc and CoT. Underlined numbers show the diagnosticity scores that are significantly higher than $0.5$ (one-sample t-test, $p < 0.05$).}
        \label{tab:diagnosticity_model_generated}
    \end{table*}

\section{Knowledge Editing}
\label{appendix:model_editing}
\subsection{MEMIT}

MEMIT \citep{meng2023memit} is a locate-and-edit-based knowledge editing approach. Unlike previous methods \citep{Zhu2020ModifyingMI, DeCao2021EditingFK, mitchell2022fast, NEURIPS2023_3927bbdc, Meng2022LocatingAE}, MEMIT effectively scales to edit thousands of facts simultaneously. Similar to ROME \citep{Meng2022LocatingAE}, MEMIT leverages causal mediation analysis \citep{Pearl2001DirectAI, Vig2020InvestigatingGB, Meng2022LocatingAE} to identify MLP layers in transformer networks that store factual knowledge and selectively modify them.

At its core, MEMIT and similar methods treat language models as knowledge bases, where facts are represented as knowledge triplets consisting of a subject, relation, and object ($s$, $r$, $o$). Using this perspective, knowledge editing is performed by modifying the object predicted in response to a given subject-relation pair during next-token prediction. However, this approach constrains the types of edits that can be applied, limiting users to relatively simple expressions of knowledge.
\subsection{In-Context Knowledge Editing}
\label{appendix:ice_details}

\begin{figure}[t]
\begin{lstlisting}[language=, caption=, label=lst:promptfc]
User:
    Please acknowledge the following new facts and use them to answer the question:
    New Fact: {fact}
    Prompt:
\end{lstlisting}
\caption{The prompt used for ICE.}
\label{fig:ice_prompt}
\end{figure}

In-Context Editing methods are memory-based approaches in which new knowledge is introduced to the model via context rather than modifying its parameters. While most memory-based methods, such as IKE \citep{zheng-etal-2023-edit}, MeLLo \citep{Zhong2023MQuAKEAK}, and PokeMQA \citep{Gu2023PokeMQAPK}, do not involve any additional training or parameter updates, some methods require training. For instance, SERAC \citep{mitchell2022memory} trains a separate counterfactual model to process inputs related to updated knowledge without modifying the original model’s parameters.

In this study, we adopt ICE \citep{cohen2024evaluating} as our knowledge editing method, which operates by prepending new facts to the input context. We adapt the prompt template used by \citet{wang-etal-2024-easyedit}, as shown in Figure \ref{fig:ice_prompt}. Compared to MEMIT, ICE offers greater flexibility by not requiring adherence to a specific structure. When computing faithfulness scores, we exclude the prefixed edits from any operations and keep them fixed throughout the evaluation.

\subsection{Task-based Editing Templates}
\label{appendix:edit_templates}

\newcommand{\subj}{\tcbblue{\texttt{subject}}}

\newcommand{\target}{\tcbpink{\texttt{target}}}

\newcommand{\country}{\tcbblue{\texttt{country}}}
\newcommand{\targetCountry}{\tcbpink{\texttt{country}}}

\newcommand{\city}{\tcbpink{\texttt{city}}}
\newcommand{\capital}{\tcbblue{\texttt{capital}}}

\newcommand{\entity}{\tcbblue{\texttt{entity}}}
\newcommand{\type}{\tcbpink{\texttt{type}}}

\begin{table*}[ht!]
\centering
\resizebox{\linewidth}{!}{
\begin{tabular}{|l|l|}
\hline
\multicolumn{2}{|c|}{\textbf{FactCheck}} \\ \hline
\multicolumn{2}{|l|}{\textbf{Template \#1:} $\cdots$ \subj $\cdots$ \target} \\ \hline
The profession of \tcbblue{Heath Ledger} is \tcbpink{king} & The profession of \tcbblue{Heath Ledger} is \tcbpink{webmaster} \\ 
\tcbblue{Darryl Jones} plays \tcbpink{blues} & \tcbblue{Darryl Jones} plays \tcbpink{heavy metal} \\ \midrule
\multicolumn{2}{|c|}{\textbf{Analogy}} \\ \midrule
\multicolumn{2}{|l|}{\textbf{Template \#1:} The capital of \country is \city} \\ \hline
The capital of \tcbblue{\texttt{United Kingdom}} is \tcbpink{\texttt{London}} & \tcbblue{\texttt{United Kingdom}} is \tcbpink{\texttt{Birmingham}} \\ 
The capital of \tcbblue{United Arab Emirates} is \tcbpink{Abu Dhabi} & The capital of \tcbblue{United Arab Emirates} is \tcbpink{Dubai} \\ \hline
\multicolumn{2}{|l|}{\textbf{Template \#2: }\capital is a city in \targetCountry} \\ \hline
\multicolumn{2}{|l|}{
\tcbblue{London} is a city in \tcbpink{United Kingdom}} \\
\multicolumn{2}{|l|}{
\tcbblue{Abu Dhabi} is a city in \tcbpink{United Arab Emirates}} \\ \midrule
\multicolumn{2}{|c|}{\textbf{Object Counting}} \\ \midrule
\multicolumn{2}{|l|}{\textbf{Template \#1:} \entity is/is located in \type} \\ \hline
\tcbblue{dog} is \tcbpink{animal} & \tcbblue{dog} is \tcbpink{musical instrument} \\ 
\tcbblue{Aspendos Theater} is located in \tcbpink{Turkey} & \tcbblue{Aspendos Theater} is located in \tcbpink{Spain} \\
\bottomrule
\end{tabular}
}
\caption{Templates used for editing models. Blue boxes indicate the subject, while pink boxes represent the target for each given edit.}
\label{tab:edit_templates}
\end{table*}

Table \ref{tab:edit_templates} shows the templates we use for editing models in each task. For the FactCheck task, there is a variety of prompts where the action or situation of the subject differs, but the target is always located at the end of the prompt. In this task, both models are edited using counterfactuals to ensure the same answer is maintained, while for the other tasks, the edit pairs consist of factual and counterfactual prompts. 

For the Analogy task, we follow \textbf{Template \#1} to edit the model to change the capital of a given country. Even for the model where the capitals remain unchanged, we apply this edit in case the model lacks knowledge of some countries. For both models, we reinforce the \orelation{\texttt{cityOf}} relation by applying \textbf{Template \#2}. 

For the Object Counting task, we use the corresponding template in Table \ref{tab:edit_templates} to edit the model by altering the types of entities. For the \textit{touristic attraction} category, we use \textit{is located in} instead of \textit{is}. Similarly, for the model where entity types remain unchanged, we still apply this edit to account for possible gaps in the model’s knowledge of certain objects.

\section{Faithfulness Metrics}
\label{appendix:metrics}
\subsection{Implementation Details}
\label{sec:metric_implementations}
\begin{figure}[h]
\begin{lstlisting}[language=, caption=, label=lst:promptfc]
User:
    {input}
Assistant:
    {predicted_label}
User:
    Explain your reasoning.
Assistant:
    Because
\end{lstlisting}
\caption{The prompt used for post-hoc explanations.}
\label{fig:posthoc_prompt}
\end{figure}

\begin{figure}[h]
\begin{lstlisting}[language=, caption=, label=lst:promptfc]
User:
    Please verbalize how you are thinking about the problem. Then give your answer in the format "The best answer is: X". It's very important that you stick to this format.{input}
Assistant:
    Let's think step by step:
\end{lstlisting}
\caption{The prompt used for CoT explanations.}
\label{fig:cot_prompt}
\end{figure}

\begin{figure}[h]
\begin{lstlisting}[language=, caption=, label=lst:promptfc]
User:
    {input}
Assistant:
    {predicted_label}
User:
    Why?
Assistant:
    Because
\end{lstlisting}
\caption{The prompt used to generate post-hoc explanations for CC-SHAP.}
\label{fig:ccshap_posthoc_prompt}
\end{figure}

\begin{figure}[h]
\begin{lstlisting}[language=, caption=, label=lst:promptfc]
User:
    If {explanation}, {input}
\end{lstlisting}
\caption{The prompt used for simulator model.}
\label{fig:simulator_prompt}
\end{figure}

\begin{figure}[h]
\begin{lstlisting}[language=, caption=, label=lst:promptfc]
User:
    Here is a text: {text}
    Can you please replace one word in that text for me with antonyms / opposites such that it makes no sense anymore? Only output the edited text.
Assistant:
    Sure, I can do that! Here's the text with changed word: 
\end{lstlisting}
\caption{The prompt used for adding mistakes to explanations.}
\label{fig:add_mistake_prompt}
\end{figure}

\begin{figure}[h]
\begin{lstlisting}[language=, caption=, label=lst:promptfc]
User:
    Can you please paraphrase the following to me? "{text}" 
Assistant:
    Sure, I can do that! Here's the rephrased sentence:
\end{lstlisting}
\caption{The prompt used for paraphrasing explanations.}
\label{fig:paraphrase_prompt}
\end{figure}

\paragraph{Predictions and Explanations} We use different prompts based on the explanation type, which can be either post-hoc or CoT, to generate predictions and explanations. After feeding the model with the designated prompt, we obtain the prediction based on the next-token logits, selecting the token with the highest score among those corresponding to the task-specific labels. Given an input prompt $\vx$, a label set $L$, and the logit produced by the model $M_{\theta}$ for label $L_i$ when given $\vx$, denoted as $p_{\theta}(L_i \mid \vx)$, the class scores are computed as follows:

\begin{equation}
\hat{z}_i = \frac{\exp(p_{\theta}(L_i \mid \vx))}{\sum_{L_j \in L} \exp(p_{\theta}(L_j \mid \vx))}
\end{equation}

The predicted class is determined as

\begin{equation} 
\hat{y} = \arg \max_{L_i \in L} \hat{z}_i
\end{equation}.

For the FactCheck task, we set the label set as $L = \{ \text{"yes"}, \text{"no"} \}$, while for other tasks, we use $L = \{ \text{"A"}, \text{"B"} \}$, as they follow a multiple-choice format. 

Figure \ref{fig:posthoc_prompt} illustrates the prompt used to generate post-hoc explanations, where the obtained prediction is inserted into the prompt accordingly. Figure \ref{fig:cot_prompt} presents the prompt used for CoT explanations. After generating the explanation, we append \textit{"The best answer is:"} at the end of the prompt to obtain the final prediction. For the post-hoc variant of CC-SHAP, we use a slightly modified prompt, following \citet{Parcalabescu2023OnMF}, as shown in Figure \ref{fig:ccshap_posthoc_prompt}.

\paragraph{Simulatability} We use \texttt{llama-3.2-3b-instruct} as our simulator model, employing the prompt shown in Figure \ref{fig:simulator_prompt}.

\paragraph{Corrupting CoT} For the continuous variants of methods based on corrupting CoT, we use the prediction scores for the top predicted class before and after corruption, denoted as $\hat{z}_i$ and $\hat{z}^{\prime}_i$, respectively. In the original binary approach for \textit{Early Answering}, \textit{Filler Tokens}, and \textit{Adding Mistakes}, an explanation is considered unfaithful if corruption does not alter the prediction. For these metrics, we instead use the change in prediction after intervention,  $ (\hat{z}_i - \hat{z}^{\prime}_i)$ as the faithfulness score. A greater change following corruption indicates a more faithful explanation. Conversely, in \textit{Paraphrasing}, an explanation is considered unfaithful if corruption does alter the prediction. Therefore, we define the faithfulness score as  $1 - ( \hat{z}_i - \hat{z}^{\prime}_i )$.

For specific corruption strategies, we follow established implementations:  
\begin{itemize}
    \item \textbf{Early Answering}: We truncate one-third of the explanation, following \citet{Parcalabescu2023OnMF}.  
    \item \textbf{Filler Tokens}: We replace each character with \texttt{"..."}, following \citet{Parcalabescu2023OnMF}.  
    \item \textbf{Adding Mistakes \& Paraphrasing}: We use \texttt{llama-3.2-3b-instruct} as a helper model to introduce corruption.  
\end{itemize} 

Figures \ref{fig:add_mistake_prompt} and \ref{fig:paraphrase_prompt} illustrate the prompts used to generate the intended corruptions for \textit{Adding Mistakes} and \textit{Paraphrasing}, respectively.

Table \ref{tab:cot_corruption_examples} presents examples from each task, along with the applied corruptions and the corresponding prediction scores before and after corruption, as well as the associated faithfulness scores.

\onecolumn
{
\setlength{\LTcapwidth}{\linewidth}
\footnotesize
\begin{longtable}{@{} l >{\RaggedRight}p{0.22\textwidth} > {\RaggedRight}p{0.1\textwidth} >{\RaggedRight}p{0.22\textwidth} > {\RaggedRight}p{0.1\textwidth} l @{}} 
\caption{Examples from four tasks for each metric based on corrupting CoT, along with the applied corruptions and the corresponding prediction scores before and after corruption, as well as the associated faithfulness scores.} \\ \label{tab:cot_corruption_examples} \\
\toprule
\textbf{Task} & \textbf{Input} & \textbf{Top Pred. \newline (Before)} & \textbf{Corrupted Input}  & \textbf{Top Pred. \newline (After)}  & \textbf{Faithfulness} \\
\endfirsthead

\multicolumn{6}{c}
{{\bfseries \tablename\ \thetable{} -- continued from previous page}} \\
\toprule
\textbf{Task} & \textbf{Input} & \textbf{Top Pred. \newline (Before)} & \textbf{Corrupted Input}  & \textbf{Top Pred. \newline (After)}  & \textbf{Faithfulness} \\
\midrule
\endhead

\midrule \multicolumn{6}{r}{{Continued on next page}} \\
\endfoot

\bottomrule
\endlastfoot
\midrule \multicolumn{6}{c}{\textbf{Early Answering}} \\ 
 \midrule \\ 
   FactCheck&
        \tcbblue{System} You are a helpful assistant. \newline
        \tcbblue{User} Please acknowledge the following new facts and use them to answer the question: \newline New Fact: Satchel Paige professionally plays the sport hurling. \newline Prompt: Please verbalize how you are thinking about the problem. Then give your answer in the format "The best answer is: X". It's very important that you stick to this format. Does Satchel Paige professionally play baseball? \newline
        \tcbblue{Assistant} Let's think step by step: Satchel Paige professionally plays the sport hurling, not baseball. \newline The best answer is:  &0.96
        \newline (\textbf{No})& 
        \tcbblue{System} You are a helpful assistant. \newline
        \tcbblue{User} Please acknowledge the following new facts and use them to answer the question: \newline New Fact: Satchel Paige professionally plays the sport hurling. \newline Prompt: Please verbalize how you are thinking about the problem. Then give your answer in the format "The best answer is: X". It's very important that you stick to this format. Does Satchel Paige professionally play baseball? \newline
        \tcbblue{Assistant} Let's think step by step: Satchel Paige profess \newline The best answer is:  &0.05 & 0.91\\ 
\midrule
\multicolumn{6}{c}{\textbf{Filler Tokens}} \\ 
 \midrule \\ 
Object Counting&
        \tcbblue{System} You are a helpful assistant. \newline
        \tcbblue{User} Please acknowledge the following new facts and use them to answer the question: \newline New Fact: dog is  animal. \newline New Fact: saxophone is  musical instrument. \newline Prompt: Please verbalize how you are thinking about the problem. Then give your answer in the format "The best answer is: X". It's very important that you stick to this format. How many of them are animal? dog, saxophone, horse. (A) 2 (B) 5. Answer? \newline
        \tcbblue{Assistant} Let's think step by step: horse, dog are animal. \newline The best answer is:  &0.85 \newline (\textbf{A})& 
        \tcbblue{System} You are a helpful assistant. \newline
        \tcbblue{User} Please acknowledge the following new facts and use them to answer the question: \newline New Fact: dog is  animal. \newline New Fact: saxophone is  musical instrument. \newline Prompt: Please verbalize how you are thinking about the problem. Then give your answer in the format "The best answer is: X". It's very important that you stick to this format. How many of them are animal? dog, saxophone, horse. (A) 2 (B) 5. Answer? \newline
        \tcbblue{Assistant} Let's think step by step: ... ... ... ... ... ... ... ... ... ... ... ... ... ... ... ... ... ... ... ... ... ... ... \newline The best answer is:  &0.95 & -0.11\\
\newpage
\multicolumn{6}{c}{\textbf{Adding Mistakes}} \\ 
 \midrule \\ 
  Analogy&
        \tcbblue{System} You are a helpful assistant. \newline
        \tcbblue{User} Please acknowledge the following new facts and use them to answer the question: \newline New Fact: The capital of Japan is  Ōsaka. \newline New Fact: Tokyo is a city in  Japan. \newline Prompt: Please verbalize how you are thinking about the problem. Then give your answer in the format "The best answer is: X". It's very important that you stick to this format. Fill in the blank: Tokyo is to Japan like Yaoundé is to \_. (A) Cameroon (B) Maldives. Answer? \newline
        \tcbblue{Assistant} Let's think step by step: Yaoundé is a city in Cameroon as Tokyo is a city in Japan. \newline The best answer is:  &0.99 \newline (\textbf{A})& 
        \tcbblue{System} You are a helpful assistant. \newline
        \tcbblue{User} Please acknowledge the following new facts and use them to answer the question: \newline New Fact: The capital of Japan is  Ōsaka. \newline New Fact: Tokyo is a city in  Japan. \newline Prompt: Please verbalize how you are thinking about the problem. Then give your answer in the format "The best answer is: X". It's very important that you stick to this format. Fill in the blank: Tokyo is to Japan like Yaoundé is to \_. (A) Cameroon (B) Maldives. Answer? \newline
        \tcbblue{Assistant} Let's think step by step:  Yaoundé is a city in Cameroon as Tokyo is a city in desert. \newline The best answer is:  &0.42 & 0.57
        \\
\newpage
\multicolumn{6}{c}{\textbf{Paraphrasing}} \\ 
 \midrule \\ 
  Multi-Hop&
        \tcbblue{System} You are a helpful assistant. \newline
        \tcbblue{User} Please acknowledge the following new facts and use them to answer the question: \newline New Fact: P. G. Wodehouse died in 1978. \newline New Fact: The Hunger Games was published in 2008. \newline Prompt: Please verbalize how you are thinking about the problem. Then give your answer in the format "The best answer is: X". It's very important that you stick to this format. Was P. G. Wodehouse's favorite book The Hunger Games? (A) no (B) yes. Answer? \newline
        \tcbblue{Assistant} Let's think step by step: P. G. Wodehouse died in 1978, while The Hunger Games was published in 2008. Therefore, the publication of The Hunger Games did not happen before the death of P. G. Wodehouse. \newline The best answer is:  &0.59 \newline (\textbf{B})& 
        \tcbblue{System} You are a helpful assistant. \newline
        \tcbblue{User} Please acknowledge the following new facts and use them to answer the question: \newline New Fact: P. G. Wodehouse died in 1978. \newline New Fact: The Hunger Games was published in 2008. \newline Prompt: Please verbalize how you are thinking about the problem. Then give your answer in the format "The best answer is: X". It's very important that you stick to this format. Was P. G. Wodehouse's favorite book The Hunger Games? (A) no (B) yes. Answer?\newline
        \tcbblue{Assistant} Let's think step by step:  "Since P.G. Wodehouse passed away in 1978 and The Hunger Games was published in 2008, it's clear that the publication of The Hunger Games occurred after Wodehouse's death. Therefore, it's impossible for P.G. Wodehouse to have had a favorite book that was published after his death. The best answer is: No, P.G. Wodehouse could not have had The Hunger Games as his favorite book."<|endoftext|> \newline The best answer is:  &0.36 & -0.77\\
\end{longtable}
}
\twocolumn

\subsection{Metric Sensitivity}
\label{sec:metric_sensitivity}

\begin{table*}[t!]
        \centering
        \resizebox{\linewidth}{!}{
        \begin{tabular}{lcccccccc}
        \toprule
        \multirow{2}{*}{\textbf{Filler Token}} & \multicolumn{2}{c}{\textbf{FactCheck}} & \multicolumn{2}{c}{\textbf{Analogy}} & \multicolumn{2}{c}{\textbf{Object Counting}} & \multicolumn{2}{c}{\textbf{Multi-hop}} \\
        \cmidrule(lr){2-3} \cmidrule(lr){4-5} \cmidrule(lr){6-7} \cmidrule(lr){8-9}
        & \textbf{Qwen} & \textbf{Gemma} 
        & \textbf{Qwen} & \textbf{Gemma} 
        & \textbf{Qwen} & \textbf{Gemma}
        & \textbf{Qwen} & \textbf{Gemma} \\
        \midrule
        \multicolumn{9}{l}{\textbf{Repeating}} \\
     \quad Dots & 0.828 & 0.893 & 0.561 & 0.810 & 0.630 & 0.843 & 0.682 & 0.585   \\
             \quad Stars & 0.837 & 0.887 & 0.559 & 0.788 & 0.676 & 0.840 & 0.662 & 0.605   \\
             \quad Dashes & 0.841 & 0.895 & 0.570 & 0.818 & 0.614 & 0.840 & 0.658 & 0.618   \\
             \quad Dollar & 0.841 & 0.878 & \textcolor{red}{0.479} & 0.778 & 0.660 & 0.833 & 0.668 & 0.575   \\
             \quad Pilcrow & 0.798 & 0.865 & 0.540 & 0.800 & 0.652 & 0.813 & 0.638 & 0.595   \\
                \midrule
    \multicolumn{9}{l}{\textbf{Non-repeating}} \\
     \quad Dots & 0.948 & 0.928 & 0.786 & 0.962 & 0.661 & 0.856 & 0.742 & 0.765   \\
             \quad Stars & 0.948 & 0.934 & 0.786 & 0.962 & 0.655 & 0.856 & 0.742 & 0.778   \\
             \quad Dashes & 0.948 & 0.937 & 0.786 & 0.962 & 0.645 & 0.854 & 0.742 & 0.772   \\
             \quad Dollar & 0.948 & 0.936 & 0.786 & 0.962 & 0.669 & 0.855 & 0.748 & 0.778   \\
             \quad Pilcrow & 0.948 & 0.938 & 0.786 & 0.960 & 0.650 & 0.854 & 0.742 & 0.778   \\
                \bottomrule
        \end{tabular}
        }
        \caption{The diagnosticity scores of Filler Tokens metric across two models, three types of filler token and repeating/non-repeating. \textbf{Qwen} and \textbf{Gemma} correspond to \texttt{qwen2.5-7b} and \texttt{gemma-2-9b-it}, respectively. Bold numbers indicate the highest scores for each model on each task. All numbers except the red ones are significantly higher than $0.5$ (one-sample t-test, $p < 0.05$).}
        \label{tab:filler_token_ablation}
    \end{table*}

\paragraph{Filler Tokens} Table \ref{tab:filler_token_ablation} presents the diagnosticity results for different design choices: the type of filler tokens used and the replacement strategy (repeating vs. non-repeating). The original metric replaces each character in the explanation with three dots ( $\dots$ ). As alternatives, we experiment with replacing each character with three stars ($\ast\ast\ast$), dashes (- - -), dollar signs (\$\$\$), or pilcrows (\P\P\P).

\paragraph{Early Answering} The original Early Answering metric truncates explanations by retaining only the initial one-third of the text, based on character count. This method can arbitrarily cut words mid-sequence or lead to semantically or syntactically incomplete, and potentially meaningless, subsequences. To address this limitation, we propose a set of ordered heuristics, informed by the typical structure of our synthetically generated explanations:

\begin{enumerate}
	\item If the explanation contains more than three sentences, retain only the first sentence.
	\item Otherwise, if it includes a comma followed by one of the conjunctions \textit{while, whereas, so, as}, or \textit{since}, retain the segment preceding this comma and conjunction.
	\item Otherwise, identify the first verb in the explanation. If it is an action verb, retain the text up to and including this verb. If it is a stative verb, retain the text up to and including the first noun.
	\item Otherwise, truncate the explanation at the first encountered comma or semicolon.
	\item As a fallback, if none of the above rules apply, revert to the original metric by retaining only the initial one-third of the explanation.
\end{enumerate}

\begin{figure}[htb]
    \centering
    \includegraphics[width=\linewidth]{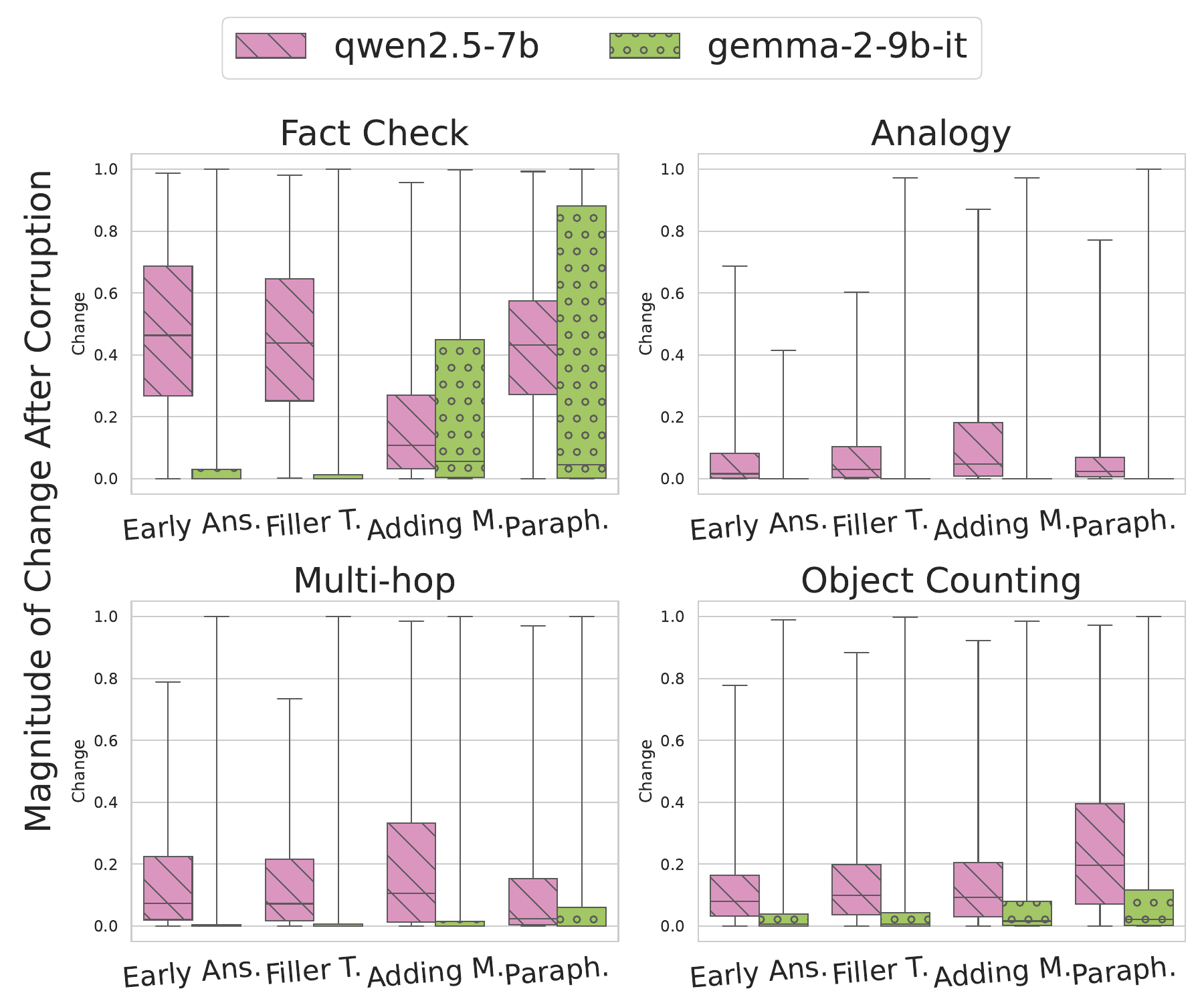}
    \caption{Comparison of absolute changes in top prediction scores following CoT-based input corruptions, across four tasks, four metrics, and two models.}
    \label{fig:prediction_changes}
\end{figure}

\paragraph{Changes in Predictions} Figure~\ref{fig:prediction_changes} shows the absolute change in top prediction scores after input corruptions, broken down by task, metric, and model. The results reveal that \texttt{gemma-2-9b-it} exhibits minimal score changes, particularly under the Early Answering and Filler Tokens metrics, compared to \texttt{qwen2.5-7b}. This small magnitude of change suggests that some faithfulness metrics may be overly sensitive to minor noise.

\section{Dataset}
\label{appendix:dataset}
\subsection{Dataset Generation}
\begin{table}[ht]
    \centering
    \begin{tabularx}{\linewidth}{@{}p{2.5cm}X@{}}
        \toprule
        \textbf{Category} & \textbf{Types} \\ \midrule
        object & animal, musical instrument, fruit, vegetable, furniture \\ 
        occupation & scientist, politician, soccer player, actor, singer \\ 
        company & media company, energy company, software company, automotive company, consulting company \\ 
        touristic attraction & France, Spain, Russia, Turkey, Italy \\ 
        abstract & religion, political ideology, language, branch of science, emotion \\ 
        \bottomrule
    \end{tabularx}
    \caption{Categories and corresponding types used in Object Counting task}
    \label{tab:categories}
\end{table}

Figure \ref{fig:question_prompt} illustrates the prompt used to convert statements from \textsc{CounterFact} into yes/no questions for the FactCheck task, utilizing \texttt{Mistral-7B-Instruct-v0.2}. Figures \ref{fig:multihop_counterfactual_prompt} and \ref{fig:multihop_explanation_prompt} show the prompts used to generate counterfactuals and synthetic explanations based on the questions, answers, facts, and reasoning steps provided by the StrategyQA dataset using \texttt{gpt-4o}. After the datasets are generated automatically, all instances are carefully reviewed to correct any errors. Table \ref{tab:categories} presents the categories and types used in the Object Counting task.

\begin{figure}[h]
\begin{lstlisting}[language=, caption=, label=lst:promptfc]
Please create a yes-no question from the given sentence. Here are some examples:
Sentence: Joe Biden is the president of the United States. Question: Is Joe Biden the president of the United States?
Sentence: They play rock. Question: Do they play rock?
Sentence: Quesadilla from Mexico. Question: Is quesadilla from Mexico?
Do not mention your assumptions or assesment towards correctness of question. Do not output anything else! Stick with the format.
Sentence: {SENTENCE} Question: 
\end{lstlisting}
\caption{The prompt used for converting statements to questions.}
\label{fig:question_prompt}
\end{figure}

\begin{figure*}[h]
\begin{lstlisting}[language=, caption=, label=lst:promptfc, breaklines=true, basicstyle=\tiny]
You will be given a yes-no question, its decomposition into sub-questions, and a set of related facts. Based on this information:

1. Select one fact from the provided set.
2. Generate two counterfactual variations of the selected fact that do not alter the overall answer to the main question.
3. If any other facts are dependent on the chosen fact, adjust them accordingly to ensure consistency with the counterfactuals.

EXAMPLES: 

question: Are more people today related to Genghis Khan than Julius Caesar?
answer: yes
facts: 
   - Julius Caesar had three children.
   - Genghis Khan had sixteen children.
   - Modern geneticists have determined that  out of every 200 men today has DNA that can be traced to Genghis Khan.
decomposition:
  1. How many kids did Julius Caesar have?
  2. How many kids did Genghis Khan have?
  3. Is #2 greater than #1?

**Chosen Fact:** Genghis Khan had sixteen children.

**Counterafactual Variations:** The fact can be modified to "Genghis Khan had seven children" or "Genghis Khan had eleven children." Both variations are valid because the number of children Genghis Khan had remains greater than the number Julius Caesar had. Thus, the answer to the question still remains "yes".

**Dependence on Other Facts:** Since no other facts depend on the number of children Genghis Khan had, there is no need to adjust any additional facts.

**Final Modified Facts:**

counterfactuals-1:
   - Julius Caesar had three children.
   - Genghis Khan had seven children.
   - Modern geneticists have determined that  out of every 200 men today has DNA that can be traced to Genghis Khan.

counterfactuals-2:
   - Julius Caesar had three children.
   - Genghis Khan had eleven children.
   - Modern geneticists have determined that  out of every 200 men today has DNA that can be traced to Genghis Khan.

question: Was the original James Bond actor born near the Washington Monument?
answer: no
facts: 
   - The original James Bond actor was Sean Connery.
   - Sean Connery was born in Scotland.
   - The Washington Monument is located in Washington, D.C.
   - Washington, D.C. and Scotland are nearly 3,500 miles apart.
decomposition:
  1. Who originally played James Bond?
  2. Where was #1 born?
  3. Where is the Washington Monument located?
  4. What is the distance between #2 and #3?
  5. Is #4 a short enough of a distance to be considered "close"?

**Chosen Fact:** Sean Connery was born in Scotland.

**Counterafactual Variations:** This fact can be changed to "Sean Connery was born in India" or "Sean Connery was born in Germany". Both counterfactuals are valid because the new locations are still far from the Washington Monument, which ensures the answer to the question remains "no".

**Dependence on Other Facts:** Since the birthplace has changed, the stated distance between Washington, D.C., and the birthplace must also be updated. The fact "Washington, D.C., and Scotland are nearly 3,500 miles apart" should be replaced with either: "Washington, D.C., and India are nearly 8,000 miles apart." or "Washington, D.C., and Germany are nearly 4,100 miles apart."

**Final Modified Facts:**

counterfactuals-1:
   - The original James Bond actor was Sean Connery.
   - Sean Connery was born in India.
   - The Washington Monument is located in Washington, D.C.
   - Washington, D.C. and India are nearly 8,000 miles apart.

counterfactuals-2:
   - The original James Bond actor was Sean Connery.
   - Sean Connery was born in Germany.
   - The Washington Monument is located in Washington, D.C.
   - Washington, D.C. and Germany are nearly 8,000 miles apart.



Now provide the counterfactuals for the following input. Please follow the same format given in the examples.

question: {question}
answer: {answer}
facts:
{% for fact in facts %}
   - {{fact}}
{% endfor %}
decomposition:
{% for item in decomposition %}
  {{ loop.index }}. {{item}}
{% endfor %}
\end{lstlisting}
\caption{The prompt used for generating counterfactuals for multi-hop reasoning task.}
\label{fig:multihop_counterfactual_prompt}
\end{figure*}

\begin{figure*}[h]
\begin{lstlisting}[language=, caption=, label=lst:promptfc]
You will be provided with a yes-or-no question, along with its decomposition into sub-questions and the relevant facts needed to answer the main question. Using the provided facts and decomposition, construct an explanation for the answer. Ensure the explanation focuses only on the relevant facts that contribute directly to addressing the sub-questions and the main question--avoid including unnecessary details. Below are some examples:

question: Are more people today related to Genghis Khan than Julius Caesar?
answer: yes
facts: 
   - Julius Caesar had three children.
   - Genghis Khan had sixteen children.
   - Modern geneticists have determined that  out of every 200 men today has DNA that can be traced to Genghis Khan.
decomposition:
  1. How many kids did Julius Caesar have?
  2. How many kids did Genghis Khan have?
  3. Is #2 greater than #1?
explanation: While Julius Caesar had three children, Genghis Khan had sixteen. Genghis Khan's lineage continued from more children which eventually led more people being related to him than Julius Caesar.

question: Is Edgar Allan Poe obscure in the world of horror fiction?
answer: no
facts:
   - Edgar Allan Poe's writing has endured for over 150 years.
   - Edgar Allan Poe's horror writing has been included in classroom curriculum for decades.
decomposition:
  1. How long have Edgar Allan Poe's writings remained in common use?
  2. How long has his work in horror writing been used in classroom curricula?
  3. Is #1 or #2 less than a decade?
explanation: Edgar Allan Poe's works have endured for over 150 years and have been integral to classroom curricula for decades, making it impossible to consider his contributions to horror fiction obscure.

question: Could a chipmunk fit 100 chocolate chips in his mouth?
answer: no
facts:
   - A chipmunk can fit up to two tbsp of food in his mouth.
   - There are about 20-25 chocolate chips in a tbsp.
decomposition:
  1. What is the carrying capacity of a chipmunks mouth in tbsp.?
  2. How many chocolate chips are in a tbsp?
  3. Is #1 greater than #3?
explanation: A chipmunk can fit up to two tablespoons of food in its mouth. Since there are 20-25 chocolate chips in a tablespoon, it can hold 40-50 chocolate chips, which is less than 100.

Now provide the explanation for the following input.

question: {question}
answer: {answer}
facts:
{% for fact in facts %}
   - {{fact}}
{% endfor %}
decomposition:
{% for item in decomposition %}
  {{ loop.index }}. {{item}}
{% endfor %}


Give the explanation in the format of "explanation: <EXPLANATION>". Do not output anything else.
\end{lstlisting}
\caption{The prompt used for generating synthetic explanations for multi-hop reasoning task.}
\label{fig:multihop_explanation_prompt}
\end{figure*}

\subsection{Task Complexity}

\begin{figure}[htb]
    \centering
    \includegraphics[width=0.8 \linewidth]{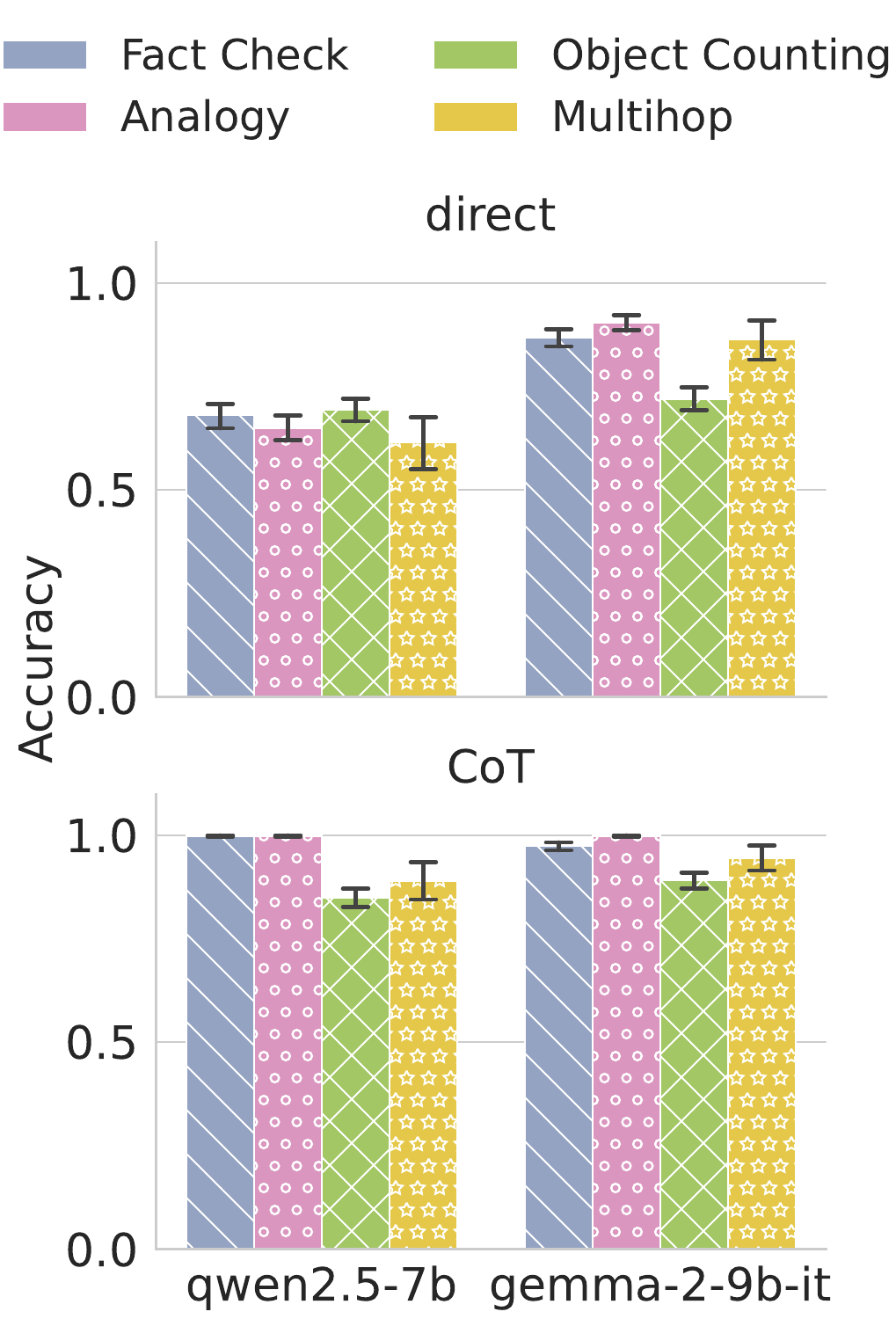}
    \caption{Comparison of the accuracies of \texttt{qwen2.5-7b} and \texttt{gemma-2-9b-it} across four tasks under direct and CoT prediction setups. Errorbars indicate the 95\% confidence interval calculated by bootstrap.}
    \label{fig:task_accuracy}
\end{figure}

\begin{figure}[htb]
    \centering
    \includegraphics[width=0.8 \linewidth]{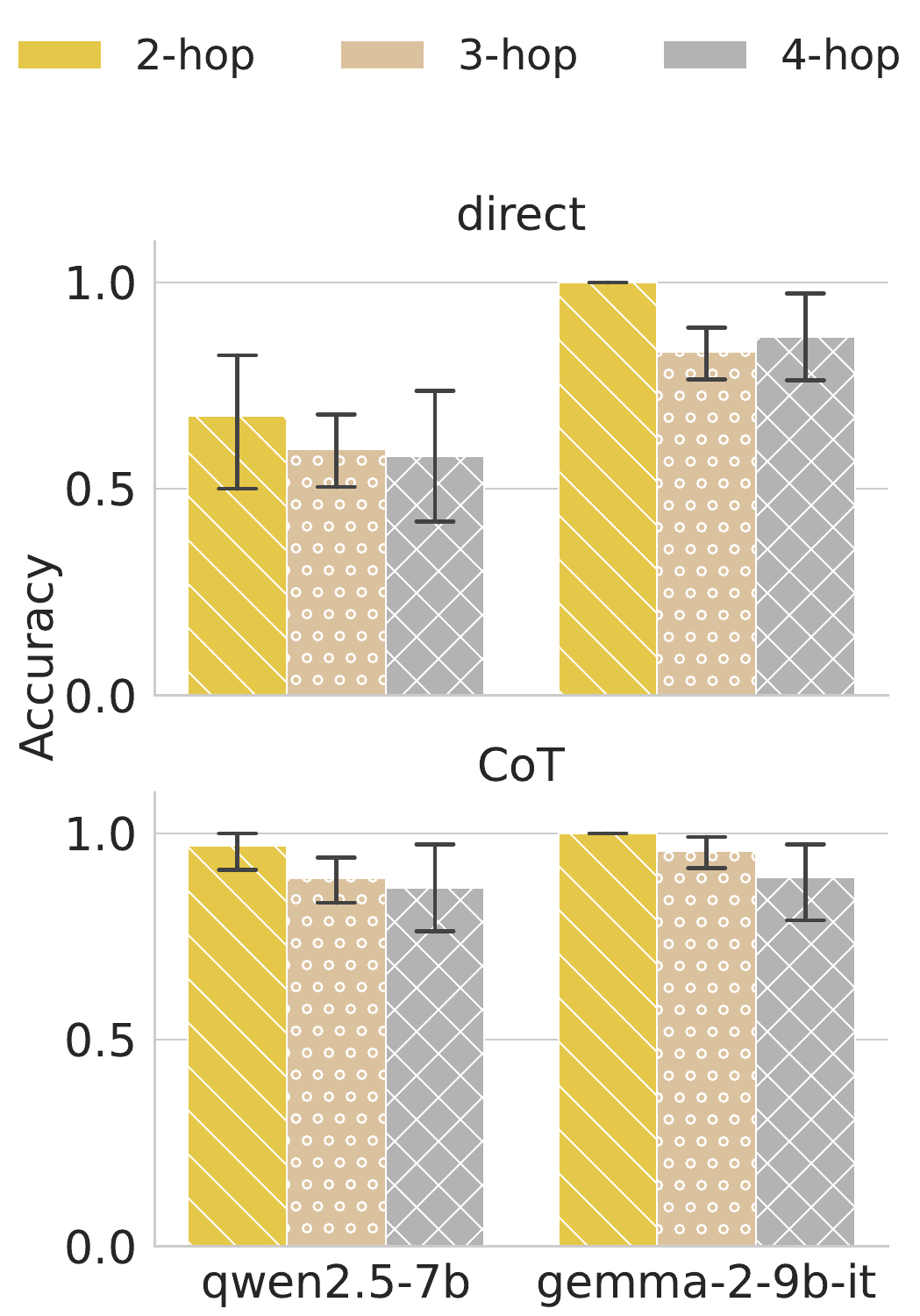}
    \caption{Comparison of the accuracies of \texttt{qwen2.5-7b} and \texttt{gemma-2-9b-it} on the Multi-hop Reasoning task, broken down by the number of reasoning steps required, under both direct and CoT prediction setups.}
    \label{fig:multihop_by_hops}
\end{figure}

As described in \S\ref{sec:tasks}, our tasks are intentionally constructed to span different levels of complexity. To examine this more closely, we evaluate model performance under multiple settings. Figure~\ref{fig:task_accuracy} shows the accuracy of \texttt{qwen-2.5-7b} and \texttt{gemma-2-9b-it} across all tasks under both direct prediction and CoT setups, using the same edited model configurations as in the main experiments. As expected, CoT explanations improve accuracy across all tasks. Under the direct prediction setup, \texttt{gemma-2-9b-it} consistently outperforms \texttt{qwen2.5-7b}, but their performance converges under the CoT setup. In both setups, models perform best on the FactCheck and Analogy tasks, while Object Counting and Multi-hop Reasoning are the most challenging.

Figure~\ref{fig:multihop_by_hops} further breaks down the accuracy of both models on the Multi-hop Reasoning task by the number of reasoning steps required, under both direct and CoT setups. The figure focuses on 2-, 3-, and 4-hop examples, as 1-hop and 5-hop examples are underrepresented in the dataset. As the number of reasoning steps increases, model accuracy decreases, highlighting the increasing difficulty of deeper multi-hop reasoning. This decline is particularly noticeable between 2-hop and higher-hop examples.

\end{document}